\documentclass[conference]{IEEEtran}
\IEEEoverridecommandlockouts

\usepackage{cite}
\usepackage{amsmath,amssymb,amsfonts}
\usepackage{graphicx}
\usepackage{xcolor}
\usepackage{booktabs}
\usepackage{adjustbox}
\usepackage{url}
\usepackage{tikz}
\usetikzlibrary{arrows.meta,positioning,calc,backgrounds}
\usepackage{pgfplots}
\pgfplotsset{compat=1.17}
\usepgfplotslibrary{groupplots}
\usepackage[hidelinks]{hyperref}

\newcommand{\acc}[2]{$#1${\scriptsize$\,\pm#2$}}
\newcommand{\dci}[3]{$#1${\scriptsize$\,[#2,#3]$}}
\newcommand{\sdci}[3]{$\mathbf{#1}${\scriptsize$\,[#2,#3]$}}
\newcommand{\U}[1]{U#1}
\newcommand{\Ug}{U448$^{+g}$}

\definecolor{cCUB}{RGB}{31,119,180}
\definecolor{cNAB}{RGB}{214,39,40}
\definecolor{cAIR}{RGB}{44,160,44}
\definecolor{cCAR}{RGB}{148,103,189}
\definecolor{cDOG}{RGB}{255,127,14}

\begin{document}

\title{When Does Looking Closer Pay Off?\\
Simple Foveated ViTs for Fine-Grained Recognition}

\author{
\IEEEauthorblockN{Boris Kriuk and Mohamed Sobhy}
\IEEEauthorblockA{
\includegraphics[height=3.5cm]{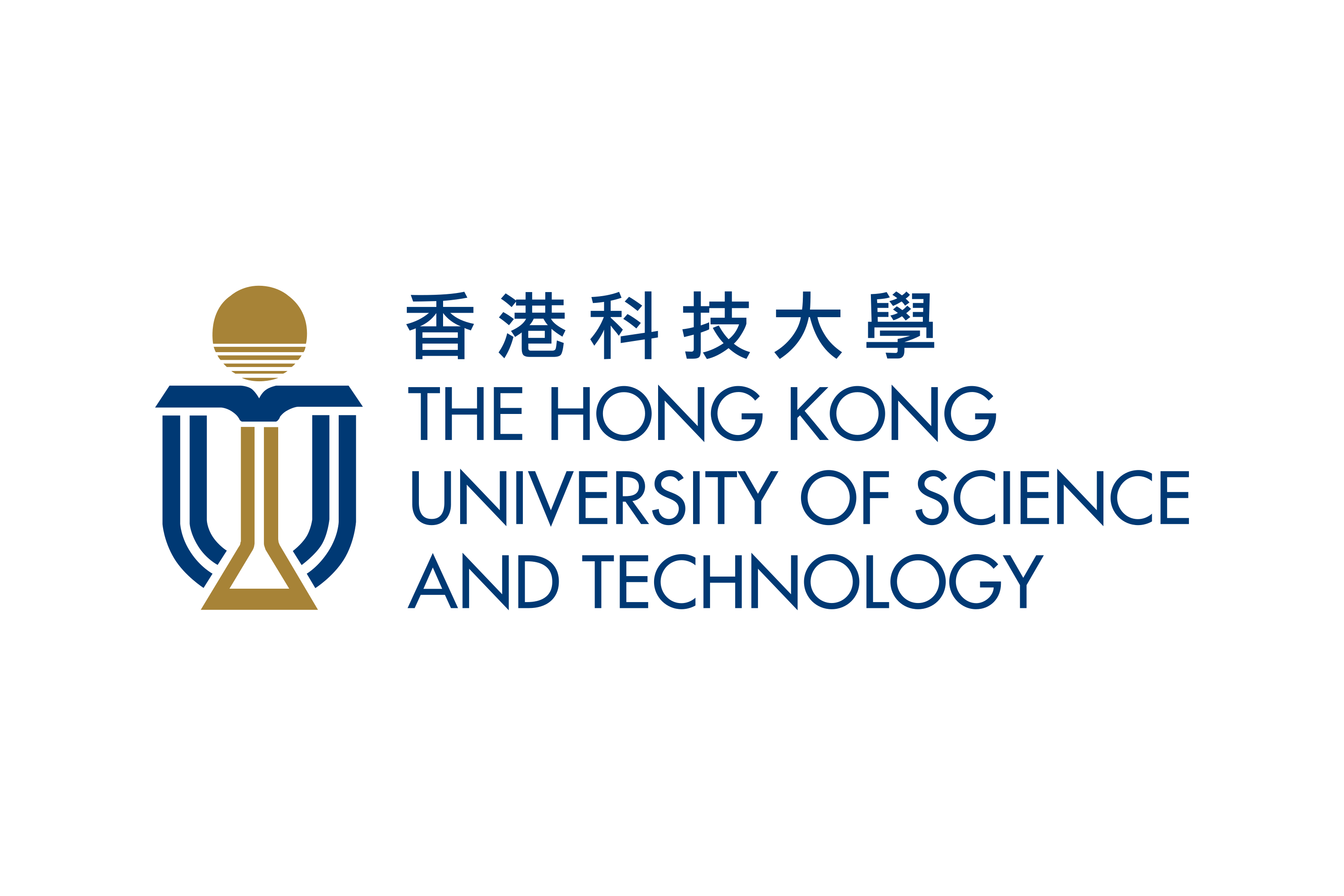} \\[1pt]
\{bkriuk, msmhmorsi\}@connect.ust.hk
}
}

\maketitle

\begin{abstract}
Fine-grained recognition benefits from input resolution, but a uniform high-resolution Vision Transformer (ViT) processes every image region at that resolution. We study selective high-resolution processing with a simple foveated ViT (FoVT). A 224-px glance passes through the backbone, whose last-block class attention selects $K$ cells. These cells are re-embedded from the 448-px image and substituted at the input for classification by the same backbone. Across five fine-grained benchmarks, two seeds, and 26 configurations per dataset under one protocol, five findings emerge. \emph{Where} to look: attention beats random selection on every dataset, by up to 2.45 points overall and 4.18 points on the smallest objects, and has higher observed accuracy than a trained center prior on four of five under the marginal interval criterion. \emph{How} to look: at equal cost, FoVT has no detected Holm-corrected deficit relative to a port of the official CF-ViT. Its small positive differences do not establish a corrected advantage in both seeds on any dataset. It requires no feature reuse, distillation loss, or selection schedule. A step-by-step decomposition finds no significant benefit from reusing contextualized coarse features; reuse costs 1--3 points when its module is trained at the head learning rate. \emph{Which} images to refine: skipping the 19--24\% most confident images leaves accuracy unchanged to two decimals. At the transformer cost of a uniform 336-px model, FoVT exceeds uniform 448-px inference or shows no significant difference on all five datasets. \emph{What} foveation buys is compute rather than accuracy: against a 448-px model trained with the same auxiliary glance loss, No accuracy difference is detected between FoVT and this model on four datasets at $0.56\times$ the cost. \emph{When} it pays off tracks selection gain and resolution headroom: the benefit is largest on NABirds and FGVC-Aircraft and is not significant on Stanford Dogs. We also document limits in budget transfer, wall-clock speed, and pre-deployment model choice.
\end{abstract}


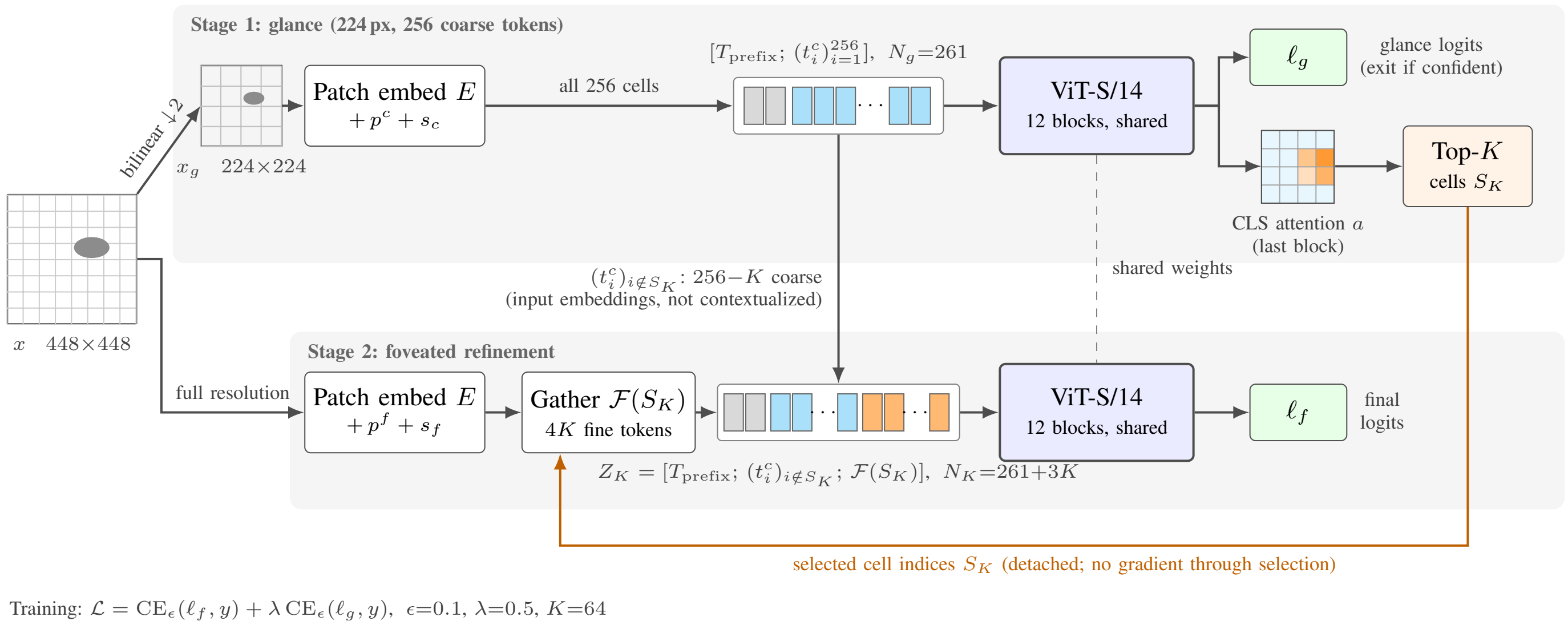
\begin{figure*}[!t]
\centering
\resizebox{\textwidth}{!}{%
\begin{tikzpicture}[
  font=\small,
  box/.style={draw=black!70, rounded corners=2pt, align=center, fill=white,
              minimum height=10mm, inner sep=3pt},
  vit/.style={box, fill=blue!8, minimum width=24mm, minimum height=12mm, thick},
  outbox/.style={box, fill=green!10, minimum width=12mm, minimum height=7mm},
  strip/.style={draw=black!50, rounded corners=1.5pt, fill=white, minimum height=7mm, inner sep=0pt},
  tokp/.style={draw=black!60, fill=black!15, minimum width=2.4mm, minimum height=4.6mm, inner sep=0pt, outer sep=0pt},
  tokc/.style={tokp, fill=cyan!30},
  tokf/.style={tokp, fill=orange!55},
  arr/.style={-{Latex[length=2mm, width=1.6mm]}, thick, black!70},
  lbl/.style={font=\scriptsize, text=black!75, align=center},
  stage/.style={font=\scriptsize\bfseries, text=black!60, anchor=north west},
]
\def\xE{4.7}
\def\xG{7.35}
\def\xS{10.2}
\def\xB{13.4}
\def\xO{15.9}
\def\xT{18.0}
\def\xR{19.2}
\begin{scope}[on background layer]
  \fill[black!4, rounded corners=4pt] (1.95,-0.75) rectangle (\xR,2.45);
  \fill[black!4, rounded corners=4pt] (3.4,-3.8)  rectangle (\xR,-1.6);
\end{scope}
\node[stage] at (2.05,2.42)  {Stage 1: glance (224\,px, 256 coarse tokens)};
\node[stage] at (3.5,-1.63)  {Stage 2: foveated refinement};
\node[draw=black!70, minimum width=16mm, minimum height=16mm, inner sep=0pt] (x) at (0.7,-0.7) {};
\begin{scope}[shift={(x.south west)}]
  \draw[black!20, step=2mm] (0,0) grid (1.6,1.6);
  \fill[black!45] (1.05,0.95) ellipse (0.22 and 0.13);
\end{scope}
\node[lbl, below=1pt of x] {$x$\quad $448{\times}448$};
\node[draw=black!70, minimum width=10mm, minimum height=10mm, inner sep=0pt] (xg) at (2.8,1.2) {};
\begin{scope}[shift={(xg.south west)}]
  \draw[black!25, step=2.5mm] (0,0) grid (1,1);
  \fill[black!45] (0.66,0.6) ellipse (0.13 and 0.08);
\end{scope}
\node[lbl, below=1pt of xg] {$x_g$\quad $224{\times}224$};
\draw[arr] (x.north east) -- (xg.west)
  node[lbl, midway, above, sloped] {bilinear $\downarrow\!2$};
\node[box, minimum width=18mm] (E1) at (\xE,1.2) {Patch embed $E$\\[-1pt]{\scriptsize$+\,p^{c}+s_c$}};
\draw[arr] (xg.east) -- (E1.west);
\node[strip, minimum width=26mm] (S1) at (\xS,1.2) {};
\begin{scope}[shift={(S1.center)}]
  \node[tokp] at (-1.05,0) {}; \node[tokp] at (-0.78,0) {};
  \node[tokc] at (-0.45,0) {}; \node[tokc] at (-0.18,0) {}; \node[tokc] at (0.09,0) {};
  \node[font=\scriptsize] at (0.42,0) {$\cdots$};
  \node[tokc] at (0.75,0) {}; \node[tokc] at (1.02,0) {};
\end{scope}
\node[lbl, above=1pt of S1] {$[T_{\mathrm{prefix}};\,(t^{c}_i)_{i=1}^{256}]$,\; $N_g{=}261$};
\draw[arr] (E1.east) -- (S1.west) node[lbl, midway, above=1pt] {all 256 cells};
\node[vit] (B1) at (\xB,1.2) {ViT-S/14\\{\scriptsize 12 blocks, shared}};
\draw[arr] (S1.east) -- (B1.west);
\node[outbox] (lg) at (\xO,1.8) {$\ell_g$};
\node[lbl, right=1pt of lg] {glance logits\\(exit if confident)};
\node[minimum width=9mm, minimum height=9mm, inner sep=0pt] (attn) at (\xO,0.45) {};
\begin{scope}[shift={(attn.south west)}]
  \fill[cyan!8] (0,0) rectangle (0.9,0.9);
  \fill[orange!25] (0.45,0.225) rectangle (0.675,0.45);
  \fill[orange!45] (0.45,0.45)  rectangle (0.675,0.675);
  \fill[orange!80] (0.675,0.45) rectangle (0.9,0.675);
  \fill[orange!60] (0.675,0.225) rectangle (0.9,0.45);
  \draw[black!35, step=2.25mm] (0,0) grid (0.9,0.9);
  \draw[black!70] (0,0) rectangle (0.9,0.9);
\end{scope}
\node[lbl, below=1pt of attn] {CLS attention $a$\\(last block)};
\draw[arr] (B1.east) -- ++(0.3,0) |- (lg.west);
\draw[arr] (B1.east) -- ++(0.3,0) |- (attn.west);
\node[box, minimum width=16mm, fill=orange!10] (topk) at (\xT,0.45) {Top-$K$\\{\scriptsize cells $S_K$}};
\draw[arr] (attn.east) -- (topk.west);
\node[box, minimum width=18mm] (E2) at (\xE,-2.6) {Patch embed $E$\\[-1pt]{\scriptsize$+\,p^{f}+s_f$}};
\draw[arr] (x.east) -- ++(0.3,0) |- (E2.west)
  node[lbl, pos=0.75, above=1pt] {full resolution};
\node[box, minimum width=17mm] (G) at (\xG,-2.6) {Gather $\mathcal{F}(S_K)$\\{\scriptsize $4K$ fine tokens}};
\draw[arr] (E2.east) -- (G.west);
\node[strip, minimum width=30mm] (S2) at (\xS,-2.6) {};
\begin{scope}[shift={(S2.center)}]
  \node[tokp] at (-1.30,0) {}; \node[tokp] at (-1.03,0) {};
  \node[tokc] at (-0.72,0) {}; \node[tokc] at (-0.45,0) {};
  \node[font=\scriptsize] at (-0.17,0) {$\cdots$};
  \node[tokc] at (0.11,0) {};
  \node[tokf] at (0.42,0) {}; \node[tokf] at (0.69,0) {};
  \node[font=\scriptsize] at (0.97,0) {$\cdots$};
  \node[tokf] at (1.25,0) {};
\end{scope}
\node[lbl, below=4pt of S2]
  {$Z_K=[T_{\mathrm{prefix}};\,(t^{c}_i)_{i\notin S_K};\,\mathcal{F}(S_K)]$,\; $N_K{=}261{+}3K$};
\draw[arr] (G.east) -- (S2.west);
\node[vit] (B2) at (\xB,-2.6) {ViT-S/14\\{\scriptsize 12 blocks, shared}};
\draw[arr] (S2.east) -- (B2.west);
\node[outbox] (lf) at (\xO,-2.6) {$\ell_f$};
\draw[arr] (B2.east) -- (lf.west);
\node[lbl, right=1pt of lf] {final\\logits};
\draw[arr] (S1.south) -- (S2.north)
  node[lbl, pos=0.62, left=2pt, align=right]
  {$(t^{c}_i)_{i\notin S_K}$: $256{-}K$ coarse\\(input embeddings, not contextualized)};
\draw[dashed, black!50] (B1.south) -- (B2.north)
  node[lbl, pos=0.55, right=2pt] {shared weights};
\draw[arr, orange!75!black] (topk.south) |- (13.0,-4.25) -| ([xshift=-6mm]G.south);
\node[lbl, text=orange!75!black] at (13.0,-4.5)
  {selected cell indices $S_K$ (detached; no gradient through selection)};
\node[lbl, anchor=west] at (-0.2,-5.05)
  {Training: $\mathcal{L}=\mathrm{CE}_{\epsilon}(\ell_f,y)+\lambda\,\mathrm{CE}_{\epsilon}(\ell_g,y)$,\;
   $\epsilon{=}0.1$, $\lambda{=}0.5$, $K{=}64$};
\end{tikzpicture}%
}
\caption{FoVT: \emph{glance, look, zoom, re-classify.} Stage~1 downsamples the $448{\times}448$ input to a $224{\times}224$ glance and runs the shared ViT-S/14 backbone to obtain glance logits $\ell_g$ and the last-block class-token attention $a$, a parameter-free saliency map. Stage~2 re-embeds the top-$K$ cells from the full-resolution image as $4K$ fine tokens (orange), which \emph{replace} the corresponding coarse input embeddings. The same backbone processes the mixed sequence $Z_K$ and produces $\ell_f$. Selection indices are detached. At inference, $K$ fixes the per-image cost, and an optional confidence gate returns $\ell_g$ and skips Stage~2 for confidently classified images.}
\label{fig:arch}
\end{figure*}

\section{Introduction}
\label{sec:intro}

Fine-grained visual categorization (FGVC) asks a model to separate classes that differ in small, spatially localized details~\cite{wei2022fgsurvey}: the bill shape of two warbler species, the grille of two sedan model years, the engine nacelles of two airliner variants~\cite{wah2011cub,vanhorn2015nabirds,maji2013aircraft,krause2013cars}. Because the evidence is small, FGVC rewards resolution~\cite{fu2017racnn,yang2018nts,hu2019wsdan}, and leading FGVC transformers are routinely run at 448~px~\cite{he2022transfg,xu2023ielt,hu2021ramstrans,wang2021ffvt}. The computational cost rises quickly. A Vision Transformer (ViT)~\cite{dosovitskiy2021vit,touvron2021deit} at 448~px processes four times as many patches as at 224~px, so the projections and MLPs cost four times as much and the attention products sixteen times as much. This cost applies to the whole frame, even when the discriminative detail occupies a single wing.

Primate vision samples at full acuity only in the fovea and uses a coarse periphery to decide where to look next~\cite{larson2009fovea,itti2001attention}. ViTs provide a spatial selection signal: the attention of the class token over patch tokens, which is known to concentrate on salient objects~\cite{abnar2020rollout,caron2021dino,chefer2021interp}. Coarse-to-fine~\cite{chen2023cfvit,wang2021dvt}, token-reduction~\cite{liang2022evit,bolya2023tome,rao2021dynamicvit,fayyaz2022ats}, and mixed-resolution~\cite{ronen2023quadformer,havtorn2023msvit} designs exploit this or similar signals. These designs can include feature reuse, learned gates, policy networks, or external scorers. Comparisons based on single runs and different protocols can obscure the effects of these components, since run-to-run variance can rival reported gains~\cite{bouthillier2021variance}. This motivates a controlled study of \emph{which} components matter and \emph{when} selective refinement is beneficial.

We study a simple foveated ViT, FoVT (Fig.~\ref{fig:arch}), under a shared evaluation protocol. FoVT (1)~\emph{glances} at a 224-px view, (2)~\emph{looks} at the glance's last-block class attention to rank the 256 cells, (3)~\emph{zooms} by re-embedding the top-$K$ cells from the 448-px image and substituting them at the input, and (4)~\emph{re-classifies} the mixed sequence with the same backbone. It has no policy network, learned gate, or selection loss, and it produces a fixed sequence length per image.

\paragraph{Study design}
We evaluate on CUB-200-2011, NABirds, FGVC-Aircraft, Stanford Cars, and Stanford Dogs~\cite{wah2011cub,vanhorn2015nabirds,maji2013aircraft,krause2013cars,khosla2011dogs} with a DINOv2 ViT-S/14 backbone~\cite{oquab2024dinov2,darcet2024registers}. We use two seeds and 26 trained configurations per dataset, for 250 training runs in total. The configurations include uniform models at four resolutions, a \emph{training-matched} uniform 448-px model that receives the same auxiliary glance loss, a port of the official CF-ViT~\cite{chen2023cfvit}, a cumulative decomposition from a reuse-based coarse-to-fine variant to FoVT, and ablations of every FoVT component. All comparisons are paired at the image level.

\paragraph{Findings}
We organize the results around five questions.
\begin{itemize}
  \item \textbf{Where to look?} The backbone's attention selects useful cells in these experiments. It beats random cells on all five datasets ($+0.72$ to $+2.45$ points, and $+1.14$ to $+4.18$ on the smallest-object tertile). It beats a trained center prior on four and a center prior swapped in at inference on all five. Averaging attention over all blocks brings no significant gain.
  \item \textbf{How to look?} At identical cost, FoVT shows no significant deficit relative to a port of the official CF-ViT: its marginal intervals exclude zero on NABirds ($+0.24$), Cars ($+0.39$), and Dogs ($+0.42$), but no advantage survives Holm correction in both seeds (Table~\ref{tab:headlineaudit}). It needs no feature reuse, distillation loss, or selection schedule. In our decomposition, fusing contextualized coarse features into the refinement pass gives no significant gain, and it costs $1$ to $3$ points when its module is trained at the head learning rate.
  \item \textbf{Which images?} A confidence gate on the glance that skips the 19--24\% most confident images leaves mean accuracy unchanged to two decimals in these evaluated runs. At the cost of a uniform 336-px model, gated FoVT exceeds uniform 448~px or shows no significant difference on all five datasets.
  \item \textbf{What does foveation buy?} Mainly compute. Against the training-matched 448-px model, No accuracy difference is detected between FoVT and this model on four datasets at $0.56\times$ the cost. Its edge over a \emph{standard} 448-px model comes mainly from the glance loss in these experiments, with an accuracy trade-off on Aircraft.
  \item \textbf{When does it pay off?} The equal-cost gain follows the selection gain and the resolution headroom beyond 336~px (Spearman $\rho=0.9$ and $\rho\approx0.87$ over the five datasets). It is largest on NABirds and Aircraft and smallest, and not significant, on Stanford Dogs.
\end{itemize}

\paragraph{Contributions}
(i)~A controlled, two-seed study of foveated refinement in fine-grained recognition, with paired bootstrap confidence intervals throughout.
(ii)~A comparison with the official CF-ViT method under the same protocol and a component-level decomposition of the coarse-to-fine design space, which show that the reported operating points can be obtained without feature reuse and that the tested reuse implementations are sensitive to their learning rate.
(iii)~Evidence that per-image confidence gating and per-image cell budgets are complementary, together with a calibration lesson: thresholds fitted on training images overshoot the budget, whereas held-out calibration approaches the target.
(iv)~A training-matched baseline that separates foveation's cost savings from the training effect of the auxiliary glance loss.
(v)~An account of when looking closer pays off: the equal-cost gain tracks selection gain and resolution headroom, is not significant on Stanford Dogs, and a validation-only recipe for choosing the model is correct in 7 of 10 seed--dataset cases.

\section{Related Work}
\label{sec:related}

\paragraph{Dynamic inference and early exits}
Dynamic networks adapt their computation to the input~\cite{han2022dynamic}. Early-exit models such as BranchyNet~\cite{teerapittayanon2016branchynet} and MSDNet~\cite{huang2018msdnet} attach intermediate classifiers and stop once a prediction is confident. RANet~\cite{yang2020ranet} and DVT~\cite{wang2021dvt} additionally escalate input resolution or token count for hard images. FoVT's confidence gate follows this tradition, but its second stage adds resolution only in selected regions rather than over the whole frame.

\paragraph{Token reduction}
DynamicViT~\cite{rao2021dynamicvit} learns to sparsify tokens hierarchically. AdaViT~\cite{meng2022adavit} learns usage policies for patches, heads, and blocks. A-ViT~\cite{yin2022avit} learns per-token halting. ATS~\cite{fayyaz2022ats} and Evo-ViT~\cite{xu2022evovit} sample or route tokens by class-attention scores. EViT~\cite{liang2022evit} keeps the most attended tokens and fuses the rest, and ToMe~\cite{bolya2023tome} merges similar tokens without training. These methods reduce computation within the network rather than selectively introducing higher-resolution input tokens.

\paragraph{Glimpses, zooming, and foveation}
Sequential glimpse models~\cite{mnih2014ram,elsayed2019saccader,wang2020gfnet} crop high-resolution patches under a learned, often reinforcement-trained, policy, which requires a second network and several passes. Differentiable alternatives warp the input toward salient regions~\cite{jaderberg2015stn,recasens2018zoom}. Foveated transformers pool peripheral tokens more coarsely~\cite{jonnalagadda2021foveater}, and PerViT~\cite{min2022pervit} imposes peripheral attention inside the network. FoVT keeps a single backbone and a single non-recurrent refinement pass.

\paragraph{Coarse-to-fine and mixed-resolution transformers}
CF-ViT~\cite{chen2023cfvit} classifies a coarse splitting, re-splits class-attention-selected regions of unconfident images, and reuses coarse-stage features in the fine stage. It is the closest relative of FoVT; we compare with a port of its official code and decompose the main design differences (Sec.~\ref{sec:cfvit}). Quadformer~\cite{ronen2023quadformer} builds a quadtree tokenization from an external saliency scorer, and MSViT~\cite{havtorn2023msvit} learns a per-region gate. Concurrently, SubViT~\cite{zhu2026subvit} (arXiv version 2) adds subtokens while retaining the original token sequence for context. Its lightweight router learns from attention supervision and avoids a separate attention forward at inference; it is evaluated on generalized category discovery. FoVT instead replaces selected coarse tokens, retains its glance pass, and studies supervised fine-grained classification. FoVT shares the single-sequence form of mixed-resolution tokenizers but takes its saliency from the backbone's own glance, without additional modules.

\paragraph{Resolution-flexible models}
FixRes~\cite{touvron2019fixres}, FlexiViT~\cite{beyer2023flexivit}, and ResFormer~\cite{tian2023resformer} let one network operate at several \emph{global} compute levels. FoVT instead controls \emph{where} resolution is spent, so the two ideas are complementary.

\paragraph{Fine-grained recognition}
Classical approaches pool second-order features~\cite{kriuk2024deep,lin2015bilinear} or localize parts with attention~\cite{zheng2017macnn}. RA-CNN~\cite{fu2017racnn}, NTS-Net~\cite{yang2018nts}, and WS-DAN~\cite{hu2019wsdan} zoom into or crop discriminative regions. TransFG~\cite{he2022transfg}, FFVT~\cite{wang2021ffvt}, RAMS-Trans~\cite{hu2021ramstrans}, IELT~\cite{xu2023ielt}, and GFT~\cite{kriuk2026gft} use attention to select, crop, or prune parts of a fixed high-resolution input, usually with larger backbones and dataset-specific recipes. We cite them as accuracy references, not compute-matched baselines. Their part-selection mechanisms are orthogonal to FoVT's input-level allocation and could be applied to its mixed-resolution sequence.

\paragraph{Evaluation methodology}
We assess all comparisons with paired tests and bootstrap intervals~\cite{dietterich1998tests,efron1993bootstrap} and report multiple seeds~\cite{bouthillier2021variance}. Because arithmetic cost is an imperfect proxy for speed~\cite{dehghani2022misnomer}, we report both MACs and measured latency.

\section{Method}
\label{sec:method}

\subsection{Mixed-Resolution Token Sequence}
\label{sec:tokens}
Let $x\in\mathbb{R}^{H\times H\times 3}$ be the input and $x_g$ its antialiased bilinear downsampling to $G\times G$, with $H=2G=448$ and patch size $P=14$. The two images yield aligned $32\times32$ fine and $16\times16$ coarse grids, so each coarse cell covers exactly a $2\times2$ group of fine patches. With a shared patch embedding $E$,
\begin{equation}
    t_i^c=E(x_g)_i+p_i^c+s_c,\qquad
    t_j^f=E(x)_j+p_j^f+s_f ,
\end{equation}
where $p^c,p^f$ are the backbone's positional embeddings resampled to the two grids and $s_c,s_f\in\mathbb{R}^D$ are zero-initialized scale embeddings. The backbone is DINOv2 ViT-S/14 with four registers~\cite{oquab2024dinov2,darcet2024registers} ($D=384$, 12 blocks). Its prefix consists of one class token and four registers. In the timm DINOv2 register-model conversion, the pretrained class positional vector is folded into the class-token initialization; registers have no positional vectors. No separate positional vector is added to either prefix during our forward passes. Uniform baselines use the same converted backbone convention. With $n=256$ coarse cells and a set $S_K$ of $K$ selected cells with fine children $\mathcal{F}(S_K)$, the second-stage sequence is
\begin{equation}
    Z_K=\big[T_{\mathrm{prefix}};\,(t_i^c)_{i\notin S_K};\,(t_j^f)_{j\in\mathcal{F}(S_K)}\big],
\end{equation}
with $n+3K$ spatial tokens. Every region is therefore represented at exactly one scale. The retained coarse tokens are the original \emph{input} embeddings, not contextualized glance outputs, and the second stage starts from fresh prefix tokens.

\subsection{Attention-Based Cell Selection}
\label{sec:selection}
The glance processes $[T_{\mathrm{prefix}};(t_i^c)_{i=1}^{n}]$ and yields logits $\ell_g$. Cell $i$ is scored by the head-averaged last-block class-token attention
\begin{equation}
    a_i=\frac{1}{A}\sum_{h=1}^{A}
    \Big[\operatorname{softmax}_{j}
    \big(q_{\mathrm{cls}}^{(h)}(k_j^{(h)})^\top/\sqrt{d_h}\big)\Big]_i ,
\end{equation}
where the softmax runs over all glance tokens and prefix entries are discarded afterwards. We set $S_K=\operatorname{TopK}(a,K)$, and the scores are detached. Registers absorb the high-norm artifact tokens that otherwise corrupt attention maps~\cite{darcet2024registers}, which supports using the raw last-block map without rollout~\cite{abnar2020rollout}. Controls replace $a$ with uniform random scores, with a center prior (the $K$ cells closest to the image center), or with attention averaged over all 12 blocks.

\subsection{Training Objective}
We fine-tune all parameters with a fixed training budget $K=64$ and supervise both predictions:
\begin{equation}
    \mathcal{L}=\operatorname{CE}_{\epsilon}(\ell_f,y)
    +\lambda\operatorname{CE}_{\epsilon}(\ell_g,y),
\end{equation}
with label smoothing $\epsilon=0.1$~\cite{szegedy2016rethinking} and $\lambda=0.5$. The glance term makes $\ell_g$ a usable early exit~\cite{teerapittayanon2016branchynet} and trains the attention that drives selection. Like deep supervision~\cite{lee2015dsn} and multi-resolution training~\cite{tian2023resformer}, it also acts as a regularizer (Sec.~\ref{sec:training}).

\subsection{Inference: Cell Budget and Confidence Gate}
\label{sec:budget}
At inference, $K$ can be changed without retraining. $K=0$ returns $\ell_g$ and skips Stage~2, and $K=256$ refines every cell. In the latter case selection is vacuous and Stage~2 reduces to a uniform 448-px pass. Separately, a \emph{confidence gate} refines an image only if the glance is unsure:
\begin{equation}
    \hat y(x)=\begin{cases}
      \arg\max \ell_f, & \max_y \operatorname{softmax}(\ell_g)_y<\eta,\\
      \arg\max \ell_g, & \text{otherwise.}
    \end{cases}
\end{equation}
For a transformer budget $B$, the refined fraction must be $f=(B-C(N_g))/C(N_K)$ (Sec.~\ref{sec:compute}). We set $\eta$ in two ways. The \emph{rank} gate refines exactly the $\operatorname{round}(fn)$ least-confident test images; it is label-free and budget-exact but needs the test pool. The \emph{calibrated} gate takes $\eta$ as the $f$-quantile of glance confidences on held-out images and is deployable per image. The gate reuses the $K=64$ checkpoint unchanged.

\subsection{Compute Accounting}
\label{sec:compute}
For $L$ blocks of width $D$ (MLP ratio four) and $N$ tokens including the prefix, we count transformer multiply--accumulates as $C(N)=L(12ND^2+2N^2D)$. With $N_g=261$ and $N_K=261+3K$, refinement costs $C(N_g)+C(N_K)$: 14.92 GMACs at $K=32$, 17.68 at $K=64$, and 23.70 at $K=128$. The glance alone costs 6.17 GMACs, and a gate with refined fraction $f$ costs $C(N_g)+fC(N_K)$. Uniform models at 224/336/392/448~px cost 6.17/15.45/22.49/31.61 GMACs. Patch embedding, heads, positional resampling, gathering, and attention extraction are excluded for all methods; measured latency is reported separately (Sec.~\ref{sec:latency}).

\subsection{Relation to CF-ViT and the Decomposition Path}
\label{sec:cfvit-method}
We compare with CF-ViT~\cite{chen2023cfvit} in two ways. First, we port its official implementation onto the shared backbone, grids, source image, and training recipe. The port keeps the official objective: cross-entropy on the fine prediction plus a KL term that distils the fine prediction into the coarse one. It also keeps the official schedule, using the full fine grid for the first two thirds of training and informative-region selection afterwards. Regions are ranked by class attention accumulated over blocks 4--12 (zero-based indices 3--11), starting from zero, with an exponential moving average of coefficient 0.99 on the previous accumulator. For feature reuse, a LayerNorm--MLP map of the coarse pass's \emph{output} features is upsampled and added to the fine tokens. The port refines a fraction $\alpha$ of the cells ($\alpha=0.25$ matches $K=64$) and can gate refinement by confidence. One change was forced by the backbone: the last reuse layer is zero-initialized and the reuse module is trained at the backbone learning rate, because with the official initialization fine-tuning from DINOv2 stalled.

Second, to examine the design choices, we build a reuse-based variant inside the FoVT code. It differs from FoVT in four ways: (a)~\emph{feature reuse}, in which a zero-initialized LayerNorm--Linear map of the glance's output features, trained at the head learning rate, is added to the coarse and fine input tokens of Stage~2; (b)~class attention uniformly averaged over all blocks, rather than the official port's exponential moving average; (c)~$\lambda=1$; and (d)~no scale embeddings. Removing (a), (b), and then (c)--(d) cumulatively turns this variant into FoVT, so each conditional step can be measured. The final step changes the loss weight and scale embeddings jointly; it does not isolate their individual effects. The official-port comparison is a comparison of complete systems.

\section{Experimental Setup}
\label{sec:setup}

\paragraph{Datasets}
CUB-200-2011 (200 classes, 5,994/5,794 train/test)~\cite{wah2011cub}, NABirds (555; 23,929/24,633)~\cite{vanhorn2015nabirds}, FGVC-Aircraft (100 variants; official trainval 6,667 / test 3,333, bottom 20-px copyright banner removed)~\cite{maji2013aircraft}, Stanford Cars (196; 8,144/8,041)~\cite{krause2013cars}, and Stanford Dogs (120; 12,000/8,580)~\cite{khosla2011dogs}. Datasets with large source images are pre-resized once to a shorter side of 672~px, which is above the 515-px evaluation resize. All five provide object boxes, which we use only for analysis: object-area tertiles of the test set and the share of selected cells on the object.

\paragraph{Protocol}
Every configuration is trained from the same timm \texttt{vit\_small\_patch14\_reg4\_dinov2.lvd142m} checkpoint with AdamW~\cite{loshchilov2019adamw}, batch size 32, BF16, one warmup epoch and cosine decay, 20 epochs (10 for NABirds), backbone/head learning rates $10^{-4}/10^{-3}$, layer-wise decay 0.85, weight decay 0.05, label smoothing 0.1, gradient clipping at 1.0, and drop-path 0.1~\cite{huang2016stochastic}. Augmentation uses 448-px random resized crops (scale $[0.3,1]$), horizontal flips, and color jitter. Evaluation resizes the shorter side to 515~px and center-crops 448~px, and uniform models receive this view downsampled. Random selection is evaluated with the random-number generator reset to seed 0 for each prediction call, producing one reproducible draw per image in the fixed evaluation order rather than averaging repeated selector draws. We evaluate the final checkpoint without early stopping or per-dataset tuning, using seeds 0 and 1 on an NVIDIA A100-SXM4-40GB (Stanford Dogs on an NVIDIA RTX 4090 Laptop GPU, and the CF-ViT port on an NVIDIA L4).

\paragraph{Configurations}
Per dataset we train uniform models at 224/336/392/448~px; FoVT with attention, random, center, or all-block selection; FoVT without scale embeddings, with $\lambda\in\{0,1\}$, with feature reuse, and trained at $K\in\{32,128\}$; FoVT at $K=256$ (a uniform 448-px model with the glance loss); the official CF-ViT port at $\alpha=0.25$; the reuse-based variant and its three decomposition steps; and, for one seed, the reuse module at backbone learning rate in FoVT and in the variant. For deployable gate calibration and validation-based model choice, we also train uniform 224/336/448 and FoVT on a stratified 90\% of the training set (split seed 20260919) and hold out the remaining 10\%.

\begin{table*}[!t]
\centering\scriptsize
\caption{Headline exact McNemar audit, conditional on each fitted run. Entries separated by a slash are seeds 0/1. Wins/losses count images correct only for A/only for B. Holm correction uses all original comparisons in each dataset--seed family; the official-port family instead spans five datasets separately for each seed and was defined during revision. Complete family membership and full-precision values accompany the analysis CSV.}
\label{tab:headlineaudit}
\begin{tabular}{llrrrr}
\toprule Comparison & Dataset & $\Delta$ (pp) & Wins:losses & Raw $p$ & Holm $p$\\
\midrule
F--random & Cub & +0.932/+1.381 & 145:91/148:68 & 0.000533/5.45e-08 & 0.0048/5.45e-07\\
F--random & Nabirds & +2.176/+2.135 & 944:408/917:391 & 3.05e-49/4.62e-49 & 3.05e-48/4.16e-48\\
F--random & Aircraft & +2.460/+2.430 & 143:61/140:59 & 9.04e-09/8.94e-09 & 9.04e-08/8.94e-08\\
F--random & Cars & +0.734/+0.709 & 124:65/126:69 & 2.13e-05/5.42e-05 & 0.000213/0.000542\\
F--random & Dogs & +0.886/+0.991 & 285:209/291:206 & 0.000724/0.000159 & 0.00724/0.00159\\
F--center & Cub & +0.535/+0.621 & 102:71/91:55 & 0.0223/0.00362 & 0.223/0.0362\\
F--center & Nabirds & +0.950/+0.873 & 560:326/559:344 & 3.45e-15/8.38e-13 & 3.8e-14/8.38e-12\\
F--center & Aircraft & +0.990/+0.960 & 91:58/89:57 & 0.00853/0.0101 & 0.0597/0.0704\\
F--center & Cars & +0.410/+0.286 & 91:58/86:63 & 0.00853/0.0711 & 0.0682/0.711\\
F--center & Dogs & +0.280/+0.256 & 201:177/212:190 & 0.237/0.295 & 1/1\\
F--matched448 & Cub & +0.017/-0.086 & 48:47/49:54 & 1/0.694 & 1/1\\
F--matched448 & Nabirds & -0.020/-0.057 & 267:272/234:248 & 0.863/0.554 & 0.863/1\\
F--matched448 & Aircraft & -0.690/-0.360 & 34:57/46:58 & 0.0206/0.281 & 0.103/1\\
F--matched448 & Cars & +0.149/-0.075 & 66:54/50:56 & 0.315/0.627 & 1/1\\
F--matched448 & Dogs & -0.128/-0.070 & 158:169/162:168 & 0.58/0.783 & 0.595/1\\
Gated F--336 & Cub & +0.466/+0.621 & 123:96/129:93 & 0.0787/0.0186 & 0.551/0.149\\
Gated F--336 & Nabirds & +0.909/+0.974 & 732:508/711:471 & 2.16e-10/3.05e-12 & 1.3e-09/2.14e-11\\
Gated F--336 & Aircraft & +2.190/+1.290 & 136:63/120:77 & 2.47e-07/0.00268 & 1.97e-06/0.0161\\
Gated F--336 & Cars & +0.597/+0.112 & 132:84/101:92 & 0.00133/0.565 & 0.00931/1\\
Gated F--336 & Dogs & +0.058/+0.280 & 252:247/272:248 & 0.858/0.313 & 1/1\\
F--official port & Cub & -0.483/-0.104 & 72:100/83:89 & 0.0392/0.703 & 0.119/1\\
F--official port & Nabirds & +0.260/+0.219 & 452:388/445:391 & 0.0297/0.0667 & 0.119/0.334\\
F--official port & Aircraft & +0.360/+0.090 & 76:64/68:65 & 0.353/0.862 & 0.353/1\\
F--official port & Cars & +0.547/+0.236 & 109:65/96:77 & 0.00106/0.171 & 0.00528/0.513\\
F--official port & Dogs & +0.361/+0.478 & 252:221/265:224 & 0.168/0.0704 & 0.335/0.334\\
\bottomrule\end{tabular}
\end{table*}

\begin{table}[!t]
\centering\scriptsize
\caption{Evaluation records for the official-port comparison. Original FoVT means are the logged main-table values; paired FoVT and CF-ViT means are computed from the aligned predictions used for their contrast. Dogs uses its original saved official-split evaluation.}
\label{tab:records}
\begin{tabular}{lrrrr}
\toprule Dataset & Original F & Paired F & CF-ViT & $\Delta$\\
\midrule
Cub & 90.11 & 90.11 & 90.40 & -0.29\\
Nabirds & 90.89 & 90.92 & 90.68 & +0.24\\
Aircraft & 91.43 & 91.51 & 91.28 & +0.23\\
Cars & 94.46 & 94.49 & 94.10 & +0.39\\
Dogs & 87.21 & 87.21 & 86.79 & +0.42\\
\bottomrule\end{tabular}
\end{table}

\paragraph{Statistics}
For each comparison we average per-image correctness over the two seeds and bootstrap over test images (5,000 resamples)~\cite{efron1993bootstrap}. These image-bootstrap intervals are conditional on the evaluated runs and do not quantify variability over future training runs. Gate predictions are formed before resampling; the bootstrap conditions on the selected test pool and does not recompute confidence ranks. We report the mean difference in points with its marginal 95\% percentile interval; bold entries indicate exclusion of zero by this unadjusted interval, not a multiplicity-controlled discovery. Unless explicitly identified as Holm-corrected, significance statements below refer only to this exploratory marginal criterion. Accuracies are reported as mean\,$\pm$\,std over seeds. Table~\ref{tab:headlineaudit} reports per-seed exact McNemar tests~\cite{mcnemar1947,dietterich1998tests} with Holm correction~\cite{holm1979}. The accompanying audit supplies all family members, integer discordances, exact-test probabilities, adjusted probabilities, and seed-direction agreement. Original families group main, training, in-house CF-ViT, decomposition, ablation, gate, and reuse-learning-rate comparisons, with correction separately within each dataset and seed. The official-port contrast is a separate family across five datasets per seed, defined during revision. Attention versus random survives correction in both seeds on all datasets; gated FoVT versus uniform 336 does so on NABirds and Aircraft. Among the official-port contrasts, only Cars seed 0 survives this correction; none establishes an advantage in both seeds. For all paired comparisons with the official CF-ViT port, we use our image-aligned evaluations of both models. We re-evaluated the saved FoVT checkpoints on CUB, NABirds, Aircraft, and Cars and used the saved official-split evaluation on Dogs. Table~\ref{tab:records} gives the accuracies underlying these contrasts alongside the logged FoVT means used in the main table. Across the four re-evaluated datasets, net correct counts differ from the logged values by at most nine images per seed. The original per-image records are unavailable, so the cause of these differences is unknown. Code, configurations, and per-image predictions will be released upon acceptance.

\section{Results}
\label{sec:results}

\begin{table*}[!t]
\centering
\caption{Top-1 accuracy (\%, mean\,$\pm$\,std over two seeds). Costs are transformer GMACs (Sec.~\ref{sec:compute}). Gated rows apply a rank confidence gate to the $K=64$ checkpoint; at both budgets their mean accuracy agrees with full refinement to two decimals. This does not assert identical predictions or identical per-seed dispersion. $^{\ast}$The training-matched model is FoVT trained at $K=256$; selection is vacuous, so its glance pass serves only as training supervision and can be skipped at inference.}
\label{tab:main}
\begin{adjustbox}{max width=\textwidth}
\small
\setlength{\tabcolsep}{4pt}
\begin{tabular}{lrccccc}
\toprule
Model & GMACs & CUB & NABirds & Aircraft & Cars & Dogs\\
\midrule
Uniform 224 & 6.17 & \acc{88.54}{0.27} & \acc{87.69}{0.09} & \acc{88.25}{0.15} & \acc{93.59}{0.10} & \acc{86.12}{0.12}\\
Uniform 336 & 15.45 & \acc{89.57}{0.06} & \acc{89.95}{0.01} & \acc{89.69}{0.36} & \acc{94.11}{0.21} & \acc{87.05}{0.01}\\
Uniform 392 & 22.49 & \acc{90.07}{0.33} & \acc{90.09}{0.33} & \acc{90.38}{0.40} & \acc{94.02}{0.06} & \acc{87.07}{0.05}\\
Uniform 448 & 31.61 & \acc{90.23}{0.00} & \acc{90.61}{0.04} & \acc{90.74}{0.49} & \acc{94.20}{0.06} & \acc{87.40}{0.27}\\
Uniform 448 + glance loss$^{\ast}$ & 31.61 & \acc{90.14}{0.24} & \acc{90.93}{0.08} & \acc{91.96}{0.51} & \acc{94.42}{0.03} & \acc{87.31}{0.11}\\
\midrule
CF-ViT (official port), $\alpha=0.25$, gated (81\% refined) & 15.45 & \acc{90.40}{0.15} & \acc{90.68}{0.09} & \acc{91.28}{0.02} & \acc{94.10}{0.06} & \acc{86.79}{0.07}\\
CF-ViT (official port), $\alpha=0.25$, all refined & 17.68 & \acc{90.40}{0.15} & \acc{90.68}{0.09} & \acc{91.28}{0.02} & \acc{94.10}{0.06} & \acc{86.79}{0.07}\\
FoVT, random cells, $K=64$ & 17.68 & \acc{88.95}{0.15} & \acc{88.74}{0.08} & \acc{88.99}{0.25} & \acc{93.74}{0.11} & \acc{86.28}{0.07}\\
\midrule
FoVT glance exit ($K=0$) & 6.17 & \acc{89.14}{0.02} & \acc{88.34}{0.06} & \acc{89.29}{0.30} & \acc{93.96}{0.07} & \acc{86.38}{0.34}\\
FoVT, $K=32$ & 14.92 & \acc{89.76}{0.18} & \acc{90.76}{0.16} & \acc{89.87}{0.02} & \acc{94.19}{0.13} & \acc{86.96}{0.02}\\
FoVT, $K=64$, gated (76\% refined) & 14.92 & \acc{90.11}{0.17} & \acc{90.89}{0.06} & \acc{91.43}{0.28} & \acc{94.46}{0.13} & \acc{87.21}{0.15}\\
FoVT, $K=64$, gated (81\% refined) & 15.45 & \acc{90.11}{0.17} & \acc{90.89}{0.06} & \acc{91.43}{0.28} & \acc{94.46}{0.13} & \acc{87.21}{0.15}\\
FoVT, $K=64$ & 17.68 & \acc{90.11}{0.17} & \acc{90.89}{0.05} & \acc{91.43}{0.28} & \acc{94.46}{0.13} & \acc{87.21}{0.15}\\ 
\bottomrule
\end{tabular}
\end{adjustbox}
\end{table*}

\begin{table*}[!t]
\centering
\caption{Paired differences (points, mean over seeds) with marginal 95\% bootstrap intervals over test images, conditional on the evaluated runs. Bold: the unadjusted interval excludes zero, not a Holm-corrected discovery. U$n$: uniform $n$-px model; \Ug: uniform 448 trained with the glance loss (training-matched); FoVT: the $K=64$ checkpoint. GMACs are given as first vs.\ second model. The first row uses the gated model at exactly the 336-px cost; its mean accuracy agrees with full refinement to two decimals; prediction identity is not inferred from this agreement. CF-ViT is the port of the official code; the gate changes no CF-ViT prediction at this budget, and the recomputed gated and full-refinement paired rows agree at the displayed precision. The last row is the deployable setting: models trained on 90\% of the training data, with the gate threshold calibrated on the held-out 10\% (realized cost 15.35--15.87 GMACs).}
\label{tab:paired}
\begin{adjustbox}{max width=\textwidth}
\footnotesize
\setlength{\tabcolsep}{3pt}
\begin{tabular}{llccccc}
\toprule
Comparison & GMACs & CUB & NABirds & Aircraft & Cars & Dogs\\
\midrule
Gated FoVT $-$ \U{336} & 15.45 / 15.45 & \sdci{+0.54}{+0.19}{+0.90} & \sdci{+0.94}{+0.74}{+1.15} & \sdci{+1.74}{+1.11}{+2.37} & \sdci{+0.35}{+0.09}{+0.60} & \dci{+0.17}{-0.20}{+0.54}\\
FoVT $-$ \U{392} & 17.68 / 22.49 & \dci{+0.04}{-0.34}{+0.41} & \sdci{+0.81}{+0.59}{+1.02} & \sdci{+1.05}{+0.45}{+1.64} & \sdci{+0.44}{+0.18}{+0.68} & \dci{+0.14}{-0.24}{+0.52}\\
FoVT $-$ \U{448} & 17.68 / 31.61 & \dci{-0.12}{-0.47}{+0.23} & \sdci{+0.29}{+0.10}{+0.48} & \sdci{+0.69}{+0.14}{+1.25} & \sdci{+0.26}{+0.02}{+0.52} & \dci{-0.18}{-0.55}{+0.20}\\
FoVT $-$ \Ug & 17.68 / 31.61 & \dci{-0.03}{-0.27}{+0.22} & \dci{-0.04}{-0.17}{+0.09} & \sdci{-0.53}{-0.98}{-0.09} & \dci{+0.04}{-0.16}{+0.22} & \dci{-0.10}{-0.40}{+0.20}\\
\Ug{} $-$ \U{448} & 31.61 / 31.61 & \dci{-0.09}{-0.44}{+0.27} & \sdci{+0.32}{+0.14}{+0.52} & \sdci{+1.22}{+0.69}{+1.76} & \dci{+0.22}{-0.01}{+0.46} & \dci{-0.08}{-0.44}{+0.29}\\
\midrule
FoVT $-$ CF-ViT ($\alpha{=}0.25$) & 17.68 / 17.68 & \dci{-0.29}{-0.60}{+0.03} & \sdci{+0.24}{+0.08}{+0.40} & \dci{+0.23}{-0.29}{+0.75} & \sdci{+0.39}{+0.14}{+0.63} & \sdci{+0.42}{+0.06}{+0.80}\\
Gated FoVT $-$ gated CF-ViT ($\alpha{=}0.25$) & 15.45 / 15.45 & \dci{-0.29}{-0.60}{+0.03} & \sdci{+0.24}{+0.08}{+0.40} & \dci{+0.23}{-0.29}{+0.75} & \sdci{+0.39}{+0.14}{+0.63} & \sdci{+0.42}{+0.06}{+0.80}\\
Glance exit $-$ \U{224} & 6.17 / 6.17 & \sdci{+0.60}{+0.24}{+0.97} & \sdci{+0.66}{+0.44}{+0.87} & \sdci{+1.04}{+0.36}{+1.68} & \sdci{+0.37}{+0.11}{+0.63} & \dci{+0.26}{-0.09}{+0.64}\\
Gated FoVT $-$ FoVT$_{K=32}$ & 14.92 / 14.92 & \sdci{+0.35}{+0.15}{+0.58} & \sdci{+0.14}{+0.03}{+0.24} & \sdci{+1.56}{+0.95}{+2.18} & \sdci{+0.27}{+0.12}{+0.43} & \sdci{+0.25}{+0.02}{+0.48}\\
\midrule
Deployable gated FoVT $-$ \U{336} & $\approx$15.5 / 15.45 & \sdci{+0.46}{+0.06}{+0.85} & \sdci{+1.23}{+1.02}{+1.44} & \sdci{+1.68}{+1.02}{+2.34} & \dci{+0.28}{-0.01}{+0.57} & \dci{+0.29}{-0.09}{+0.66}\\
\bottomrule
\end{tabular}
\end{adjustbox}
\end{table*}

\subsection{Does Foveation Pay Off?}
\label{sec:main}
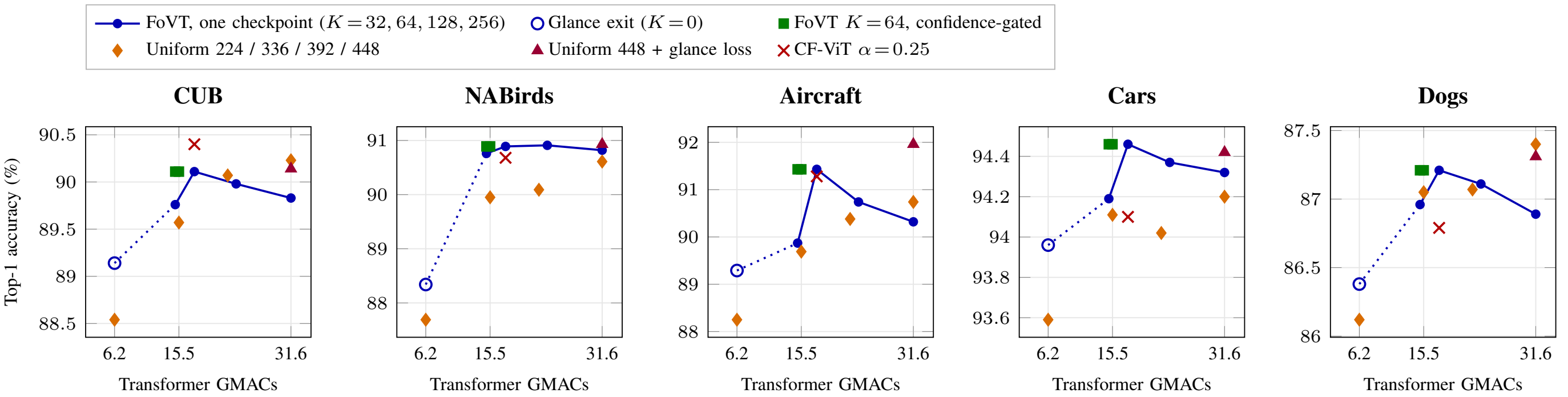
\begin{figure*}[!t]
\centering
\resizebox{\textwidth}{!}{%
\begin{tikzpicture}
\begin{groupplot}[
  group style={group size=5 by 1, horizontal sep=1.15cm},
  width=4.6cm, height=4.4cm,
  xmin=2, xmax=34.5, xtick={6.17,15.45,31.61},
  xticklabels={6.2,15.5,31.6},
  xlabel={Transformer GMACs},
  grid=major, grid style={black!10},
  tick label style={font=\scriptsize}, label style={font=\scriptsize},
  title style={font=\small\bfseries, yshift=-1mm, text height=1.6ex, text depth=0.5ex},
  yticklabel style={/pgf/number format/fixed, /pgf/number format/precision=1},
]
\nextgroupplot[title={CUB}, ylabel={Top-1 accuracy (\%)},
  legend style={font=\scriptsize, at={(0,1.27)}, anchor=south west, legend columns=3, draw=black!30,
    /tikz/every even column/.append style={column sep=8pt}},
  legend cell align=left]
\addplot[blue!70!black, dotted, thick, forget plot] coordinates {(6.17,89.14) (14.92,89.76)};
\addplot[blue!70!black, thick, mark=*, mark size=1.4pt] coordinates {(14.92,89.76) (17.68,90.11) (23.70,89.98) (31.61,89.83)};
\addlegendentry{FoVT, one checkpoint ($K\!=\!32,64,128,256$)}
\addplot[blue!70!black, only marks, mark=o, mark size=2.2pt, thick] coordinates {(6.17,89.14)};
\addlegendentry{Glance exit ($K\!=\!0$)}
\addplot[green!50!black, only marks, mark=square*, mark size=1.8pt] coordinates {(14.92,90.11) (15.45,90.11)};
\addlegendentry{FoVT $K\!=\!64$, confidence-gated}
\addplot[orange!85!black, only marks, mark=diamond*, mark size=2.4pt] coordinates {(6.17,88.54) (15.45,89.57) (22.49,90.07) (31.61,90.23)};
\addlegendentry{Uniform 224 / 336 / 392 / 448}
\addplot[purple!80!black, only marks, mark=triangle*, mark size=2.4pt] coordinates {(31.61,90.14)};
\addlegendentry{Uniform 448 + glance loss}
\addplot[red!70!black, only marks, mark=x, mark size=3pt, thick] coordinates {(17.68,90.40)};
\addlegendentry{CF-ViT $\alpha\!=\!0.25$}
\nextgroupplot[title={NABirds}]
\addplot[blue!70!black, dotted, thick] coordinates {(6.17,88.34) (14.92,90.76)};
\addplot[blue!70!black, thick, mark=*, mark size=1.4pt] coordinates {(14.92,90.76) (17.68,90.89) (23.70,90.91) (31.61,90.82)};
\addplot[blue!70!black, only marks, mark=o, mark size=2.2pt, thick] coordinates {(6.17,88.34)};
\addplot[green!50!black, only marks, mark=square*, mark size=1.8pt] coordinates {(14.92,90.89) (15.45,90.89)};
\addplot[orange!85!black, only marks, mark=diamond*, mark size=2.4pt] coordinates {(6.17,87.69) (15.45,89.95) (22.49,90.09) (31.61,90.61)};
\addplot[purple!80!black, only marks, mark=triangle*, mark size=2.4pt] coordinates {(31.61,90.93)};
\addplot[red!70!black, only marks, mark=x, mark size=3pt, thick] coordinates {(17.68,90.68)};
\nextgroupplot[title={Aircraft}]
\addplot[blue!70!black, dotted, thick] coordinates {(6.17,89.29) (14.92,89.87)};
\addplot[blue!70!black, thick, mark=*, mark size=1.4pt] coordinates {(14.92,89.87) (17.68,91.43) (23.70,90.74) (31.61,90.32)};
\addplot[blue!70!black, only marks, mark=o, mark size=2.2pt, thick] coordinates {(6.17,89.29)};
\addplot[green!50!black, only marks, mark=square*, mark size=1.8pt] coordinates {(14.92,91.43) (15.45,91.43)};
\addplot[orange!85!black, only marks, mark=diamond*, mark size=2.4pt] coordinates {(6.17,88.25) (15.45,89.69) (22.49,90.38) (31.61,90.74)};
\addplot[purple!80!black, only marks, mark=triangle*, mark size=2.4pt] coordinates {(31.61,91.96)};
\addplot[red!70!black, only marks, mark=x, mark size=3pt, thick] coordinates {(17.68,91.28)};
\nextgroupplot[title={Cars}, yticklabel style={/pgf/number format/fixed, /pgf/number format/precision=2}]
\addplot[blue!70!black, dotted, thick] coordinates {(6.17,93.96) (14.92,94.19)};
\addplot[blue!70!black, thick, mark=*, mark size=1.4pt] coordinates {(14.92,94.19) (17.68,94.46) (23.70,94.37) (31.61,94.32)};
\addplot[blue!70!black, only marks, mark=o, mark size=2.2pt, thick] coordinates {(6.17,93.96)};
\addplot[green!50!black, only marks, mark=square*, mark size=1.8pt] coordinates {(14.92,94.46) (15.45,94.46)};
\addplot[orange!85!black, only marks, mark=diamond*, mark size=2.4pt] coordinates {(6.17,93.59) (15.45,94.11) (22.49,94.02) (31.61,94.20)};
\addplot[purple!80!black, only marks, mark=triangle*, mark size=2.4pt] coordinates {(31.61,94.42)};
\addplot[red!70!black, only marks, mark=x, mark size=3pt, thick] coordinates {(17.68,94.10)};
\nextgroupplot[title={Dogs}]
\addplot[blue!70!black, dotted, thick] coordinates {(6.17,86.38) (14.92,86.96)};
\addplot[blue!70!black, thick, mark=*, mark size=1.4pt] coordinates {(14.92,86.96) (17.68,87.21) (23.70,87.11) (31.61,86.89)};
\addplot[blue!70!black, only marks, mark=o, mark size=2.2pt, thick] coordinates {(6.17,86.38)};
\addplot[green!50!black, only marks, mark=square*, mark size=1.8pt] coordinates {(14.92,87.21) (15.45,87.21)};
\addplot[orange!85!black, only marks, mark=diamond*, mark size=2.4pt] coordinates {(6.17,86.12) (15.45,87.05) (22.49,87.07) (31.61,87.40)};
\addplot[purple!80!black, only marks, mark=triangle*, mark size=2.4pt] coordinates {(31.61,87.31)};
\addplot[red!70!black, only marks, mark=x, mark size=3pt, thick] coordinates {(17.68,86.79)};
\end{groupplot}
\end{tikzpicture}%
}
\caption{Accuracy--compute operating points (mean of two seeds). Blue: one FoVT checkpoint evaluated at several budgets; $K=256$ is plotted at 31.61 GMACs because its glance pass can be skipped. Green squares: the same checkpoint at $K=64$ with a confidence gate at the costs of $K=32$ and of uniform 336 (14.92 and 15.45 GMACs; the two markers nearly coincide at this scale). Orange: separately trained uniform models. Purple: uniform 448 trained with the glance loss. Red crosses: the official CF-ViT port. Gating dominates lowering $K$ on every dataset. Vertical ranges differ across panels.}
\label{fig:frontier}
\end{figure*}

Tables~\ref{tab:main} and~\ref{tab:paired} and Fig.~\ref{fig:frontier} summarize the main comparison. At $K=64$ (17.68 GMACs, $0.56\times$ uniform 448), FoVT shows no significant deficit relative to any standard uniform model on any dataset. It significantly exceeds uniform 448 on NABirds ($+0.29$), Aircraft ($+0.69$), and Cars ($+0.26$) with no significant difference on CUB or Dogs. It also exceeds uniform 392 on NABirds, Aircraft, and Cars. On NABirds the gain concentrates on small objects: on the smallest-object tertile FoVT exceeds uniform 392 by $+1.48$ points (interval $[+1.08,+1.88]$), compared with $+0.81$ overall.

With the confidence gate, the same accuracy is available at the cost of a uniform 336-px model. Gated FoVT beats uniform 336 at identical cost on four datasets ($+0.35$ to $+1.74$) with no significant difference on Dogs. At $0.49\times$ the cost of uniform 448, FoVT exceeds uniform 448 or shows no significant difference on all five datasets. In the deployable setting with held-out calibration, the gain over uniform 336 remains positive on all five datasets and significant on CUB, NABirds, and Aircraft. At the same 6.17 GMACs, the glance exit beats a dedicated uniform 224-px model by $+0.37$ to $+1.04$ points on four datasets with no significant difference on Dogs ($+0.26$).

The last two rows of the first block of Table~\ref{tab:paired} qualify these gains. A uniform 448-px model trained with the same auxiliary glance loss improves over the standard 448-px model on NABirds ($+0.32$) and Aircraft ($+1.22$). No accuracy difference is detected between FoVT and this training-matched model on four datasets and trails it by $0.53$ points on Aircraft. These comparisons attribute FoVT's advantage over standard high-resolution training mainly to the glance loss. Foveation reduces the cost to $0.56\times$, with no significant accuracy difference on four datasets (Sec.~\ref{sec:training}).

\begin{table*}[!t]
\centering
\caption{FoVT ablations as paired differences FoVT $-$ variant (points; positive means the FoVT choice is better). ``Trained'' variants are separate models; ``swap'' variants change only the selector of the trained FoVT checkpoint at inference. All rows cost 17.68 GMACs, so these are strictly compute-matched comparisons. Bold: 95\% interval excludes zero.}
\label{tab:ablation}
\begin{adjustbox}{max width=\textwidth}
\footnotesize
\setlength{\tabcolsep}{3.5pt}
\begin{tabular}{lccccc}
\toprule
Variant & CUB & NABirds & Aircraft & Cars & Dogs\\
\midrule
\multicolumn{6}{l}{\emph{Where to look}}\\
Random cells (trained) & \sdci{+1.16}{+0.77}{+1.54} & \sdci{+2.16}{+1.93}{+2.38} & \sdci{+2.45}{+1.77}{+3.14} & \sdci{+0.72}{+0.47}{+0.97} & \sdci{+0.94}{+0.56}{+1.31}\\
\quad smallest-object tertile & \sdci{+1.99}{+1.24}{+2.77} & \sdci{+4.18}{+3.72}{+4.66} & \sdci{+2.66}{+1.35}{+3.96} & \sdci{+1.14}{+0.65}{+1.62} & \sdci{+1.61}{+0.89}{+2.33}\\
Center prior (trained) & \sdci{+0.58}{+0.26}{+0.88} & \sdci{+0.91}{+0.73}{+1.09} & \sdci{+0.98}{+0.42}{+1.53} & \sdci{+0.35}{+0.12}{+0.56} & \dci{+0.27}{-0.06}{+0.61}\\
All-block attention (trained) & \dci{-0.06}{-0.28}{+0.16} & \dci{-0.08}{-0.20}{+0.04} & \dci{+0.06}{-0.35}{+0.47} & \sdci{+0.20}{+0.02}{+0.38} & \dci{+0.10}{-0.17}{+0.37}\\
Random cells (swap) & \sdci{+1.25}{+0.89}{+1.61} & \sdci{+2.43}{+2.20}{+2.65} & \sdci{+3.26}{+2.58}{+3.96} & \sdci{+0.52}{+0.29}{+0.75} & \sdci{+1.09}{+0.75}{+1.42}\\
Center prior (swap) & \sdci{+0.91}{+0.60}{+1.23} & \sdci{+1.08}{+0.92}{+1.24} & \sdci{+2.16}{+1.55}{+2.76} & \sdci{+0.74}{+0.51}{+0.97} & \sdci{+0.97}{+0.65}{+1.29}\\
All-block attention (swap) & \dci{+0.07}{-0.09}{+0.22} & \dci{-0.01}{-0.09}{+0.06} & \dci{+0.02}{-0.29}{+0.32} & \dci{+0.07}{-0.06}{+0.20} & \dci{+0.15}{-0.05}{+0.36}\\
\midrule
\multicolumn{6}{l}{\emph{Tokens and loss}}\\
No scale embeddings & \dci{+0.12}{-0.09}{+0.33} & \dci{+0.01}{-0.10}{+0.13} & \dci{+0.00}{-0.39}{+0.38} & \dci{+0.02}{-0.13}{+0.18} & \dci{+0.20}{-0.06}{+0.48}\\
No glance loss ($\lambda=0$) & \dci{+0.02}{-0.28}{+0.31} & \sdci{+0.42}{+0.26}{+0.58} & \sdci{+1.10}{+0.59}{+1.59} & \sdci{+0.41}{+0.19}{+0.63} & \sdci{+0.44}{+0.12}{+0.77}\\
Glance loss $\lambda=1$ & \dci{+0.11}{-0.11}{+0.34} & \dci{+0.05}{-0.08}{+0.18} & \dci{+0.24}{-0.15}{+0.63} & \dci{+0.07}{-0.09}{+0.23} & \dci{+0.10}{-0.16}{+0.37}\\
\bottomrule
\end{tabular}
\end{adjustbox}
\end{table*}

\subsection{Where to Look}
\label{sec:where}
The glance's last-block attention selects useful cells across the tested datasets (Table~\ref{tab:ablation}). Trained with random cells at the same sequence length, the model loses $0.72$ to $2.45$ points, with larger losses on the smallest-object tertile. Trained with a center prior, it loses $0.35$ to $0.98$ points on four datasets but only a non-significant $0.27$ on Dogs, whose subjects fill most of the frame. Swapping the selector of a trained attention model to random or center cells at inference costs $0.52$ to $3.26$ points. This supports image-specific attention over a dataset-level spatial prior. Uniform all-block averaging, which differs from the exponential moving average used by CF-ViT~\cite{chen2023cfvit}, gives no significant gain on any dataset and is $0.20$ points worse on Cars. It also adds latency (0.64--0.66 vs.\ 0.56~ms/image), so we keep the simpler single-block rule in this setting.

The selected cells land on the annotated object at high rates: 87\%, 83\%, 97\%, 99\%, and 93\% on CUB, NABirds, Aircraft, Cars, and Dogs. Random selection reaches 52\%, 48\%, 55\%, 73\%, and 71\%, the object's mean area fraction. The selected cells cover 47\%, 52\%, 47\%, 36\%, and 38\% of the object area. Consistent with the object-centric attention of self-supervised ViTs~\cite{caron2021dino}, selection without any box supervision thus localizes reliably even on Dogs, where looking closer brings no significant gain at equal cost (Sec.~\ref{sec:dogs}).

\begin{table*}[!t]
\centering
\caption{From a reuse-based coarse-to-fine variant to FoVT, one design choice at a time (all 17.68 GMACs, every image refined). Top: accuracy (\%, two seeds unless marked). Bottom: paired differences between consecutive rows or as indicated. Variant (0) trains its reuse module at the head learning rate; bb.\ LR: reuse module trained at the backbone learning rate. The port of the official CF-ViT is listed for reference. $^{\ddagger}$Single seed (seed 0); intervals bootstrap over test images only. Bold: interval excludes zero.}
\label{tab:decomp}
\begin{adjustbox}{max width=\textwidth}
\footnotesize
\setlength{\tabcolsep}{3.5pt}
\begin{tabular}{lccccc}
\toprule
Configuration & CUB & NABirds & Aircraft & Cars & Dogs\\
\midrule
(0) Reuse-based variant (reuse, all-block, $\lambda{=}1$) & \acc{88.74}{0.35} & \acc{89.15}{0.11} & \acc{88.34}{0.11} & \acc{93.30}{0.15} & \acc{85.90}{0.24}\\
(1) $-$ feature reuse & \acc{89.95}{0.04} & \acc{90.76}{0.04} & \acc{91.36}{0.08} & \acc{94.33}{0.07} & \acc{86.85}{0.06}\\
(2) $+$ last-block attention & \acc{89.89}{0.23} & \acc{90.76}{0.11} & \acc{91.30}{0.17} & \acc{94.28}{0.05} & \acc{87.00}{0.10}\\
(3) $+$ $\lambda{=}0.5$, scale emb.\ ($\equiv$\,FoVT) & \acc{90.15}{0.18} & \acc{90.98}{0.09} & \acc{91.45}{0.21} & \acc{94.37}{0.02} & \acc{87.27}{0.02}\\
FoVT (reference impl.) & \acc{90.11}{0.17} & \acc{90.89}{0.05} & \acc{91.43}{0.28} & \acc{94.46}{0.13} & \acc{87.21}{0.15}\\
FoVT $+$ reuse & \acc{88.70}{0.04} & \acc{89.73}{0.32} & \acc{88.13}{0.19} & \acc{93.35}{0.11} & \acc{85.90}{0.24}\\
FoVT $+$ reuse (bb.\ LR)$^{\ddagger}$ & 89.78 & 90.50 & 91.12 & 94.32 & 86.85\\
(0) with reuse at bb.\ LR$^{\ddagger}$ & 89.66 & 90.33 & 91.51 & 94.07 & 86.67\\
CF-ViT (official port) & \acc{90.40}{0.15} & \acc{90.68}{0.09} & \acc{91.28}{0.02} & \acc{94.10}{0.06} & \acc{86.79}{0.07}\\
\midrule
(1)$-$(0): remove reuse & \sdci{+1.21}{+0.77}{+1.65} & \sdci{+1.61}{+1.39}{+1.82} & \sdci{+3.02}{+2.28}{+3.75} & \sdci{+1.03}{+0.72}{+1.32} & \sdci{+0.94}{+0.52}{+1.39}\\
(2)$-$(1): last-block attention & \dci{-0.05}{-0.26}{+0.16} & \dci{+0.00}{-0.11}{+0.12} & \dci{-0.06}{-0.44}{+0.33} & \dci{-0.05}{-0.24}{+0.14} & \dci{+0.16}{-0.11}{+0.43}\\
(3)$-$(2): $\lambda$ and scale emb. & \sdci{+0.26}{+0.03}{+0.49} & \sdci{+0.22}{+0.09}{+0.35} & \dci{+0.15}{-0.27}{+0.57} & \dci{+0.09}{-0.10}{+0.27} & \dci{+0.26}{-0.02}{+0.55}\\
FoVT$-$(3): sanity check & \dci{-0.04}{-0.20}{+0.11} & \dci{-0.09}{-0.18}{+0.01} & \dci{-0.02}{-0.35}{+0.30} & \dci{+0.09}{-0.04}{+0.23} & \dci{-0.05}{-0.27}{+0.19}\\
(FoVT $+$ reuse) $-$ FoVT & \sdci{-1.41}{-1.84}{-0.98} & \sdci{-1.17}{-1.37}{-0.97} & \sdci{-3.30}{-4.07}{-2.58} & \sdci{-1.11}{-1.41}{-0.80} & \sdci{-1.31}{-1.78}{-0.86}\\
(FoVT $+$ reuse, bb.\ LR) $-$ FoVT$^{\ddagger}$ & \dci{-0.21}{-0.60}{+0.19} & \sdci{-0.36}{-0.58}{-0.14} & \dci{-0.51}{-1.20}{+0.18} & \dci{-0.24}{-0.53}{+0.06} & \dci{-0.26}{-0.73}{+0.20}\\
(1) $-$ ((0) with reuse at bb.\ LR)$^{\ddagger}$ & \dci{+0.26}{-0.16}{+0.69} & \sdci{+0.40}{+0.18}{+0.62} & \dci{-0.09}{-0.75}{+0.57} & \dci{+0.21}{-0.09}{+0.51} & \dci{+0.14}{-0.34}{+0.62}\\
FoVT $-$ CF-ViT (official port) & \dci{-0.29}{-0.60}{+0.03} & \sdci{+0.24}{+0.08}{+0.40} & \dci{+0.23}{-0.29}{+0.75} & \sdci{+0.39}{+0.14}{+0.63} & \sdci{+0.42}{+0.06}{+0.80}\\
\bottomrule
\end{tabular}
\end{adjustbox}
\end{table*}

\subsection{How to Look: FoVT versus CF-ViT}
\label{sec:cfvit}
At identical cost, FoVT and the port of the official CF-ViT are close on every dataset (Table~\ref{tab:paired}). FoVT is ahead on NABirds ($+0.24$), Cars ($+0.39$), and Dogs ($+0.42$). The difference on Aircraft ($+0.23$) is not significant, and CF-ViT is ahead on CUB by $0.29$ points, also not significantly. Refinement helps both models: FoVT's refined prediction exceeds its own glance by $+0.50$ to $+2.55$ points, and CF-ViT's exceeds its coarse prediction by $+0.35$ to $+2.37$. Both also tolerate the confidence gate. Refining only the 81\% least-confident images changes no CF-ViT prediction, and recomputation confirms that the paired comparison at 15.45 GMACs agrees with the one at 17.68 to the displayed precision. In a single-seed check under an earlier protocol, training CF-ViT at $\alpha=0.5$ (23.70 GMACs with every image refined) changed its accuracy by $-0.06$ to $+0.11$ points on four datasets and by $+0.43$ on Dogs. FoVT thus achieves comparable accuracy to the CF-ViT port with fewer components: it has no reuse module, no distillation loss, and no selection schedule, and it ranks cells with a single attention map.

Table~\ref{tab:decomp} tests successive changes to the reuse-based variant of Sec.~\ref{sec:cfvit-method}. Switching from uniform all-block averaging to last-block attention produces no significant change, and FoVT's loss weight and scale embeddings add at most $0.26$ points. The final configuration reproduces our FoVT implementation within $0.09$ points. Feature reuse is the sensitive component. With its module trained at the head learning rate, it costs $0.94$ to $3.02$ points in the variant and $1.11$ to $3.30$ points when added to FoVT. At the backbone learning rate most of this loss disappears (single seed): the variant comes within $0.40$ points of its counterpart without reuse, and FoVT with reuse within $0.51$ points of FoVT, with a significant deficit remaining only on NABirds. The official port, whose reuse module is zero-initialized and trained at the backbone learning rate, fits this pattern. Feature reuse was not needed to obtain the reported operating points. We detected no benefit for the tested reuse implementations and learning rates; the large penalties occurred at the head learning rate. These null results do not establish equivalence or rule out a useful reuse effect under other settings. High-resolution input embeddings suffice for the second pass in these experiments.

\subsection{Which Images: Confidence Gating}
\label{sec:gate}
Refinement can be skipped for confident images without a measurable accuracy loss at the tested budgets. Refining only the 76\% or 81\% least-confident images (14.92 and 15.45 GMACs) reproduces the fully refined accuracy to two decimals on all five datasets (Table~\ref{tab:main}). As Fig.~\ref{fig:frontier} and Table~\ref{tab:paired} show, at a fixed budget it is better to gate a $K=64$ model than to lower $K$: the gated model beats $K=32$ at identical cost on every dataset ($+0.14$ to $+1.56$). An alternative, refining half of the images with $K=128$ at the same 14.92 GMACs, is comparable on four datasets but worse on Aircraft (90.79\%). Gating and per-image cell budgets thus control two different axes: whether an image needs a closer look, and how much of it.

\begin{table}[!t]
\centering
\caption{Calibrating the gate for a target refined fraction of 81\% (15.45 GMACs). Thresholds from training images overshoot, because fine-tuned models are more confident on their training data (mean glance confidence, train vs.\ test). Thresholds from a held-out 10\% split approach the target.}
\label{tab:gatecal}
\begin{adjustbox}{max width=\columnwidth}
\footnotesize
\setlength{\tabcolsep}{3pt}
\begin{tabular}{lccc}
\toprule
 & Train-calibrated & Held-out-calibrated & Confidence\\
Dataset & frac.\ / GMACs & frac.\ / GMACs & train / test\\
\midrule
CUB & 0.93 / 16.87 & 0.82 / 15.66 & 0.90 / 0.74\\
NABirds & 0.91 / 16.60 & 0.84 / 15.87 & 0.87 / 0.73\\
Aircraft & 0.91 / 16.65 & 0.80 / 15.35 & 0.89 / 0.78\\
Cars & 0.90 / 16.51 & 0.81 / 15.44 & 0.89 / 0.79\\
Dogs & 0.91 / 16.65 & 0.82 / 15.56 & 0.90 / 0.74\\
\bottomrule
\end{tabular}
\end{adjustbox}
\end{table}

How the threshold is set matters for deployment (Table~\ref{tab:gatecal}). Calibrated on training images, the threshold refines 90--93\% of test images instead of 81\%, overshooting the budget by 7--9\%. This overshoot is consistent with the higher confidence of the fine-tuned glance on training images (0.87--0.90) than on test images (0.73--0.79), an instance of the overconfidence of modern networks on the data they were fit to~\cite{guo2017calibration}. Calibrated on a held-out 10\% split, the realized fraction is 80--84\%, within 2.7\% of the target cost. The largest remaining mismatch is on NABirds, whose validation images are slightly more confident than its test images (0.75 vs.\ 0.72). This held-out setting yields the deployable comparisons in the last row of Table~\ref{tab:paired}.

\begin{table}[!t]
\centering
\caption{Budget transfer (\%, mean of two seeds). Upper block: one $K=64$ checkpoint at several inference budgets; $K=256$ can be served at 31.61 GMACs. Lower block: separate checkpoints trained at the evaluated budget, with paired differences to the single checkpoint at the same $K$ shown in parentheses where significant. Bold: best budget per dataset for the single checkpoint.}
\label{tab:budgets}
\begin{adjustbox}{max width=\columnwidth}
\footnotesize
\setlength{\tabcolsep}{2.6pt}
\begin{tabular}{rrccccc}
\toprule
$K$ & GMACs & CUB & NAB. & Air. & Cars & Dogs\\
\midrule
0 & 6.17 & 89.14 & 88.34 & 89.29 & 93.96 & 86.38\\
32 & 14.92 & 89.76 & 90.76 & 89.87 & 94.19 & 86.96\\
64 & 17.68 & \textbf{90.11} & 90.89 & \textbf{91.43} & \textbf{94.46} & \textbf{87.21}\\
128 & 23.70 & 89.98 & \textbf{90.91} & 90.74 & 94.37 & 87.11\\
256 & 31.61 & 89.83 & 90.82 & 90.32 & 94.32 & 86.89\\
\midrule
32$^{\text{tr}}$ & 14.92 & 89.95 & 90.86 & 90.47 & 94.06 & 87.16\\
128$^{\text{tr}}$ & 23.70 & 90.21 & 90.99 & 91.78 & 94.38 & 87.37\\
 & & {\scriptsize($+0.22$)} & & {\scriptsize($+1.04$)} & & \\
\bottomrule
\end{tabular}
\end{adjustbox}
\end{table}

\subsection{How Many Cells: Budget Transfer}
\label{sec:k}
A single checkpoint transfers well across budgets on four datasets (Table~\ref{tab:budgets}). On CUB, NABirds, Cars, and Dogs, every $K\geq32$ stays within 0.35 points of the best budget, and checkpoints trained at $K=32$ or $K=128$ show no significant difference from the single checkpoint at those budgets, except on CUB at $K=128$ ($+0.22$). On NABirds, $K=32$ already matches uniform 448 at $0.47\times$ its cost ($+0.15$, interval $[-0.04,+0.35]$). Aircraft is the exception. Its accuracy peaks sharply at the training budget: $K=16$ (86.70\%) falls below the glance exit, $K=32$ barely exceeds it, and $K=128$ loses 0.69 points. A checkpoint trained at $K=128$ recovers $+1.04$ points there. Inference-time budget flexibility is therefore not free on every task, much like the train--test resolution discrepancy of uniform models~\cite{touvron2019fixres}. When compute must be lowered, the gate (Sec.~\ref{sec:gate}) preserves accuracy more reliably in these experiments; when it can be raised, training at the target budget, or across budgets as in FlexiViT~\cite{beyer2023flexivit}, is advisable on tasks like Aircraft.

\subsection{What the Glance Loss Contributes}
\label{sec:training}
The auxiliary glance loss plays three roles. First, it makes the glance a usable exit: without it the glance exit is less accurate (58.0\% on Aircraft, 91.9\% on Cars), whereas with it the exit beats a dedicated 224-px model on four datasets, with no significant difference on Dogs (Table~\ref{tab:paired}). Second, it improves the refined prediction itself. Removing it ($\lambda=0$) costs $0.41$ to $1.10$ points on NABirds, Aircraft, Cars, and Dogs with no significant effect on CUB, while no difference is detected between $\lambda=1$ and $\lambda=0.5$ (Table~\ref{tab:ablation}). Without the glance loss, FoVT still beats uniform 392 on NABirds ($+0.38$) but no longer exceeds uniform 448 anywhere. Third, it can improve full-resolution training: the same loss added to a uniform 448-px model raises it by $+0.32$ on NABirds and $+1.22$ on Aircraft, in line with the benefits of deep supervision~\cite{lee2015dsn} and multi-resolution training~\cite{tian2023resformer}.

These numbers decompose FoVT's advantage over a standard uniform 448-px model into a foveation term (FoVT $-$ training-matched 448) and a training term (training-matched 448 $-$ standard 448). On CUB the split is $-0.03$ and $-0.09$; on NABirds $-0.04$ and $+0.32$; on Aircraft $-0.53$ and $+1.22$; on Cars $+0.04$ and $+0.22$; on Dogs $-0.10$ and $-0.08$. Wherever FoVT beats standard 448-px training, the training term accounts for most of the gain. The foveation term is near zero on four datasets. FoVT shows no significant difference from the training-matched model at $0.56\times$ the cost, or $0.49\times$ with the gate. On Aircraft, refining a quarter of the cells yields a deficit of 0.53 points. Fine details distributed over the airframe may explain this result.

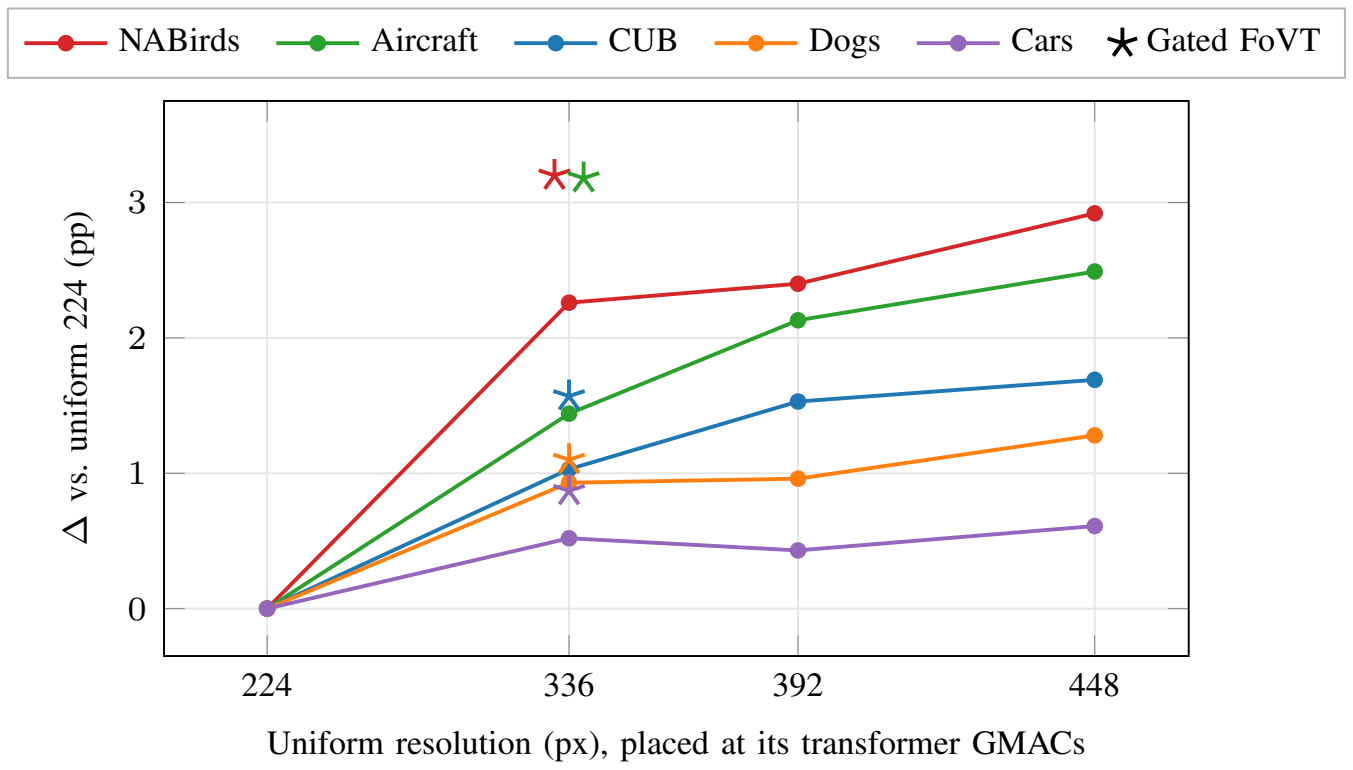
\begin{figure}[!t]
\centering
\resizebox{\columnwidth}{!}{%
\begin{tikzpicture}
\begin{axis}[
  width=9cm, height=5.6cm,
  xmin=3, xmax=34.5, ymin=-0.35, ymax=3.75,
  xtick={6.17,15.45,22.49,31.61}, xticklabels={224,336,392,448},
  xlabel={Uniform resolution (px), placed at its transformer GMACs},
  ylabel={$\Delta$ vs.\ uniform 224 (pp)},
  grid=major, grid style={black!10},
  tick label style={font=\scriptsize}, label style={font=\scriptsize},
  legend style={font=\scriptsize, at={(0.5,1.04)}, anchor=south, legend columns=6, draw=black!30,
    /tikz/every even column/.append style={column sep=5pt}},
  legend cell align=left,
]
\addplot[cNAB, thick, mark=*, mark size=1.3pt] coordinates {(6.17,0) (15.45,2.26) (22.49,2.40) (31.61,2.92)};
\addlegendentry{NABirds}
\addplot[cAIR, thick, mark=*, mark size=1.3pt] coordinates {(6.17,0) (15.45,1.44) (22.49,2.13) (31.61,2.49)};
\addlegendentry{Aircraft}
\addplot[cCUB, thick, mark=*, mark size=1.3pt] coordinates {(6.17,0) (15.45,1.03) (22.49,1.53) (31.61,1.69)};
\addlegendentry{CUB}
\addplot[cDOG, thick, mark=*, mark size=1.3pt] coordinates {(6.17,0) (15.45,0.93) (22.49,0.96) (31.61,1.28)};
\addlegendentry{Dogs}
\addplot[cCAR, thick, mark=*, mark size=1.3pt] coordinates {(6.17,0) (15.45,0.52) (22.49,0.43) (31.61,0.61)};
\addlegendentry{Cars}
\addlegendimage{black, only marks, mark=star, mark size=3.2pt, thick}
\addlegendentry{Gated FoVT}
\addplot[cNAB, only marks, mark=star, mark size=3.4pt, thick, xshift=-3pt, forget plot] coordinates {(15.45,3.20)};
\addplot[cAIR, only marks, mark=star, mark size=3.4pt, thick, xshift=3pt, forget plot] coordinates {(15.45,3.18)};
\addplot[cCUB, only marks, mark=star, mark size=3.4pt, thick, forget plot] coordinates {(15.45,1.57)};
\addplot[cDOG, only marks, mark=star, mark size=3.4pt, thick, forget plot] coordinates {(15.45,1.10)};
\addplot[cCAR, only marks, mark=star, mark size=3.4pt, thick, forget plot] coordinates {(15.45,0.87)};
\end{axis}
\end{tikzpicture}%
}
\caption{Uniform-resolution scaling relative to each dataset's 224-px model (mean of two seeds). Stars: gated FoVT at the 336-px cost; the NABirds and Aircraft stars ($+3.20$ and $+3.18$) are offset horizontally for legibility. On NABirds and Aircraft, accuracy keeps rising up to 448~px and gated FoVT lies above the entire uniform curve. On Dogs, most of the uniform gain is already reached at 336~px, and gated FoVT adds little to it.}
\label{fig:resscale}
\end{figure}

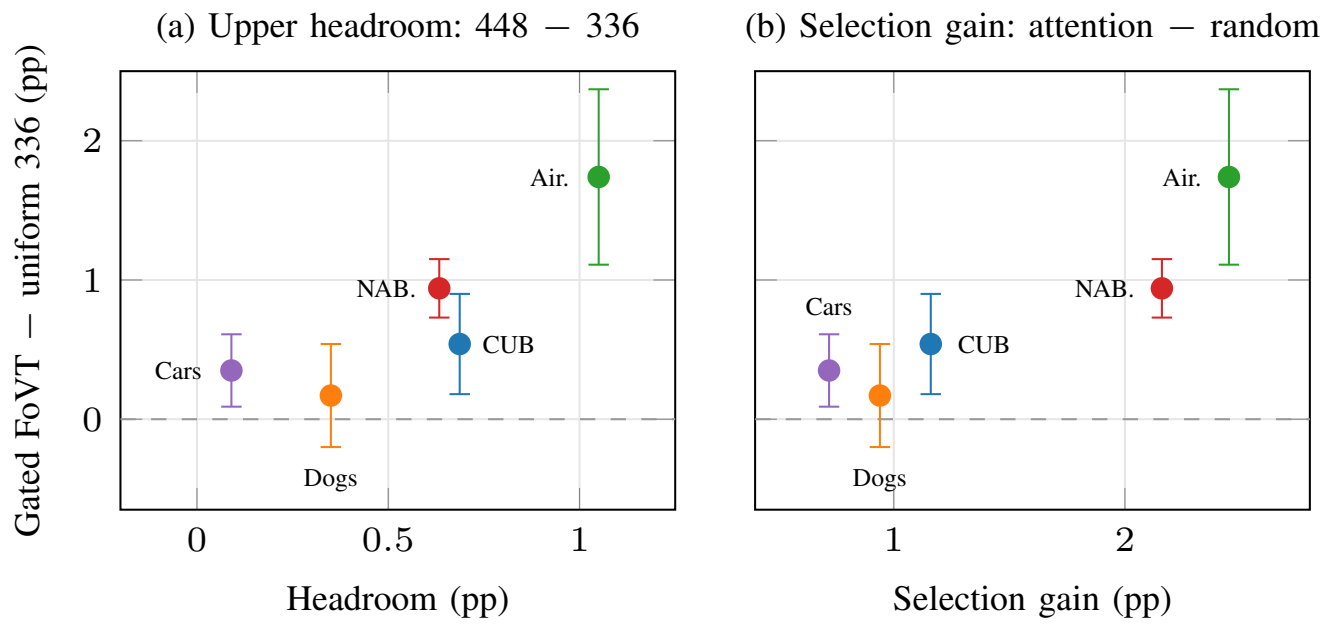
\begin{figure}[!t]
\centering
\resizebox{\columnwidth}{!}{%
\begin{tikzpicture}
\begin{groupplot}[
  group style={group size=2 by 1, horizontal sep=0.55cm, yticklabels at=edge left},
  width=5.4cm, height=4.6cm,
  ymin=-0.65, ymax=2.5,
  grid=major, grid style={black!10},
  tick label style={font=\scriptsize}, label style={font=\scriptsize},
  title style={font=\scriptsize, yshift=-1.5mm},
]
\nextgroupplot[title={(a) Upper headroom: 448 $-$ 336},
  xmin=-0.2, xmax=1.25, xlabel={Headroom (pp)}, ylabel={Gated FoVT $-$ uniform 336 (pp)}]
\addplot[black!40, dashed, domain=-0.2:1.25, samples=2, forget plot] {0};
\addplot[only marks, mark=*, mark size=2pt, cCUB, xshift=2pt, error bars/.cd, y dir=both, y explicit] coordinates {(0.66,0.54) +- (0,0.36)};
\addplot[only marks, mark=*, mark size=2pt, cNAB, xshift=-2pt, error bars/.cd, y dir=both, y explicit] coordinates {(0.66,0.94) +- (0,0.21)};
\addplot[only marks, mark=*, mark size=2pt, cAIR, error bars/.cd, y dir=both, y explicit] coordinates {(1.05,1.74) +- (0,0.63)};
\addplot[only marks, mark=*, mark size=2pt, cCAR, error bars/.cd, y dir=both, y explicit] coordinates {(0.09,0.35) +- (0,0.26)};
\addplot[only marks, mark=*, mark size=2pt, cDOG, error bars/.cd, y dir=both, y explicit] coordinates {(0.35,0.17) +- (0,0.37)};
\node[font=\tiny, anchor=west] at (axis cs:0.70,0.54) {CUB};
\node[font=\tiny, anchor=east] at (axis cs:0.62,0.94) {NAB.};
\node[font=\tiny, anchor=east] at (axis cs:1.02,1.74) {Air.};
\node[font=\tiny, anchor=east] at (axis cs:0.06,0.35) {Cars};
\node[font=\tiny, anchor=north] at (axis cs:0.35,-0.23) {Dogs};
\nextgroupplot[title={(b) Selection gain: attention $-$ random},
  xmin=0.4, xmax=2.8, xlabel={Selection gain (pp)}]
\addplot[black!40, dashed, domain=0.4:2.8, samples=2, forget plot] {0};
\addplot[only marks, mark=*, mark size=2pt, cCUB, error bars/.cd, y dir=both, y explicit] coordinates {(1.16,0.54) +- (0,0.36)};
\addplot[only marks, mark=*, mark size=2pt, cNAB, error bars/.cd, y dir=both, y explicit] coordinates {(2.16,0.94) +- (0,0.21)};
\addplot[only marks, mark=*, mark size=2pt, cAIR, error bars/.cd, y dir=both, y explicit] coordinates {(2.45,1.74) +- (0,0.63)};
\addplot[only marks, mark=*, mark size=2pt, cCAR, error bars/.cd, y dir=both, y explicit] coordinates {(0.72,0.35) +- (0,0.26)};
\addplot[only marks, mark=*, mark size=2pt, cDOG, error bars/.cd, y dir=both, y explicit] coordinates {(0.94,0.17) +- (0,0.37)};
\node[font=\tiny, anchor=west] at (axis cs:1.20,0.54) {CUB};
\node[font=\tiny, anchor=east] at (axis cs:2.12,0.94) {NAB.};
\node[font=\tiny, anchor=east] at (axis cs:2.41,1.74) {Air.};
\node[font=\tiny, anchor=south] at (axis cs:0.72,0.62) {Cars};
\node[font=\tiny, anchor=north] at (axis cs:0.94,-0.23) {Dogs};
\end{groupplot}
\end{tikzpicture}%
}
\caption{When does looking closer pay off? Equal-cost gain of gated FoVT over uniform 336 (with 95\% intervals) against (a)~the resolution headroom beyond 336~px and (b)~the gain of attention over random selection. Both orderings largely match the ordering of the gains (Spearman $\rho\approx0.87$ and $\rho=0.9$); the main mismatch is that both rank Dogs above Cars. CUB and NABirds have the same headroom (0.66) and are offset horizontally in (a) for legibility. With five datasets these correlations are descriptive.}
\label{fig:when}
\end{figure}

\subsection{When Does Looking Closer Pay Off?}
\label{sec:when}
Fig.~\ref{fig:resscale} shows where each dataset's resolution benefit lies. NABirds and Aircraft keep improving up to 448~px, CUB improves mainly between 336 and 392~px, Cars barely responds to resolution, and Dogs is flat from 336 to 392~px, with a non-significant $+0.35$ points at 448~px. Fig.~\ref{fig:when} relates the equal-cost gain of gated FoVT to two dataset statistics. The \emph{upper headroom} (uniform 448 $-$ uniform 336: 1.05, 0.66, 0.66, 0.09, and 0.35 on Aircraft, NABirds, CUB, Cars, and Dogs) largely reproduces the ordering of the gains ($\rho\approx0.87$). Because both quantities share the 336-px baseline, we treat this as suggestive. The \emph{selection gain} (attention $-$ random: 2.45, 2.16, 1.16, 0.72, 0.94) does not use the uniform-model measurements and orders the gains equally well ($\rho=0.9$). However, selection gain and equal-cost gain both contain the FoVT accuracy term when gating preserves accuracy. Their association is therefore also outcome-coupled, rather than independent corroboration; both correlations are descriptive. Both statistics misorder only Cars and Dogs, the two datasets with the smallest gains ($+0.35$ and $+0.17$). Object size alone is a weaker predictor ($\rho=-0.6$ with mean box area): Cars has the largest objects (0.73 of the crop) yet gains, partly through the glance loss (Sec.~\ref{sec:training}).

These results suggest that looking closer pays off when the discriminative evidence is small, so that \emph{where} to look carries information, and when that evidence still benefits from resolution beyond a moderate input size. Bird and aircraft identification satisfy both conditions. We hypothesize similar behavior for species identification in citizen-science and camera-trap imagery~\cite{vanhorn2018inat}, product recognition on cluttered shelves, and defect inspection, where most pixels are context rather than evidence.

\paragraph{Can the decision be made before deployment?}
We explored a validation-based recipe: choose FoVT if upper headroom on the held-out 10\% split is positive, otherwise choose uniform 336 (Table~\ref{tab:choiceaudit}). We treat this rule as exploratory because its specification before inspection of the test outcomes is not documented. Using the observed full-refinement FoVT-minus-uniform-336 test difference consistently, the recipe selects the higher-accuracy model in 7 of 10 seed--dataset cases. Always choosing FoVT does so in 10 of 10, and always choosing uniform 336 in 0 of 10. Mean observed accuracy regret is 0.17 points for the recipe, zero for always choosing FoVT, and 0.75 points for always choosing uniform 336. The case record therefore does not show an advantage over the constant FoVT choice. These are accuracy-only comparisons; engineering and latency costs remain separate (Sec.~\ref{sec:latency}).

\begin{table}[!t]
\centering\scriptsize
\caption{Exploratory validation choice. Headroom is validation uniform 448 minus uniform 336. Test differences compare full-refinement FoVT with uniform 336. Correct means selecting the observed higher accuracy, not an established population-level superiority. Regret is the observed accuracy loss relative to the better of these two choices.}
\label{tab:choiceaudit}
\begin{tabular}{lrrrrr}
\toprule Dataset & Seed & Headroom & Choice & Test $\Delta$ & Regret\\
\midrule
Cub & 0 & -1.50 & U & +0.47 & 0.47\\
Cub & 1 & -0.33 & U & +0.62 & 0.62\\
Nabirds & 0 & +1.58 & F & +0.91 & 0.00\\
Nabirds & 1 & +0.58 & F & +0.97 & 0.00\\
Aircraft & 0 & +0.29 & F & +2.19 & 0.00\\
Aircraft & 1 & +0.29 & F & +1.29 & 0.00\\
Cars & 0 & -0.63 & U & +0.60 & 0.60\\
Cars & 1 & +0.51 & F & +0.11 & 0.00\\
Dogs & 0 & +0.25 & F & +0.06 & 0.00\\
Dogs & 1 & +0.08 & F & +0.28 & 0.00\\
\bottomrule\end{tabular}
\end{table}

\begin{figure*}[!t]
\centering
\begin{minipage}[t]{0.49\textwidth}
\centering\small NABirds\par
\includegraphics[width=\linewidth]{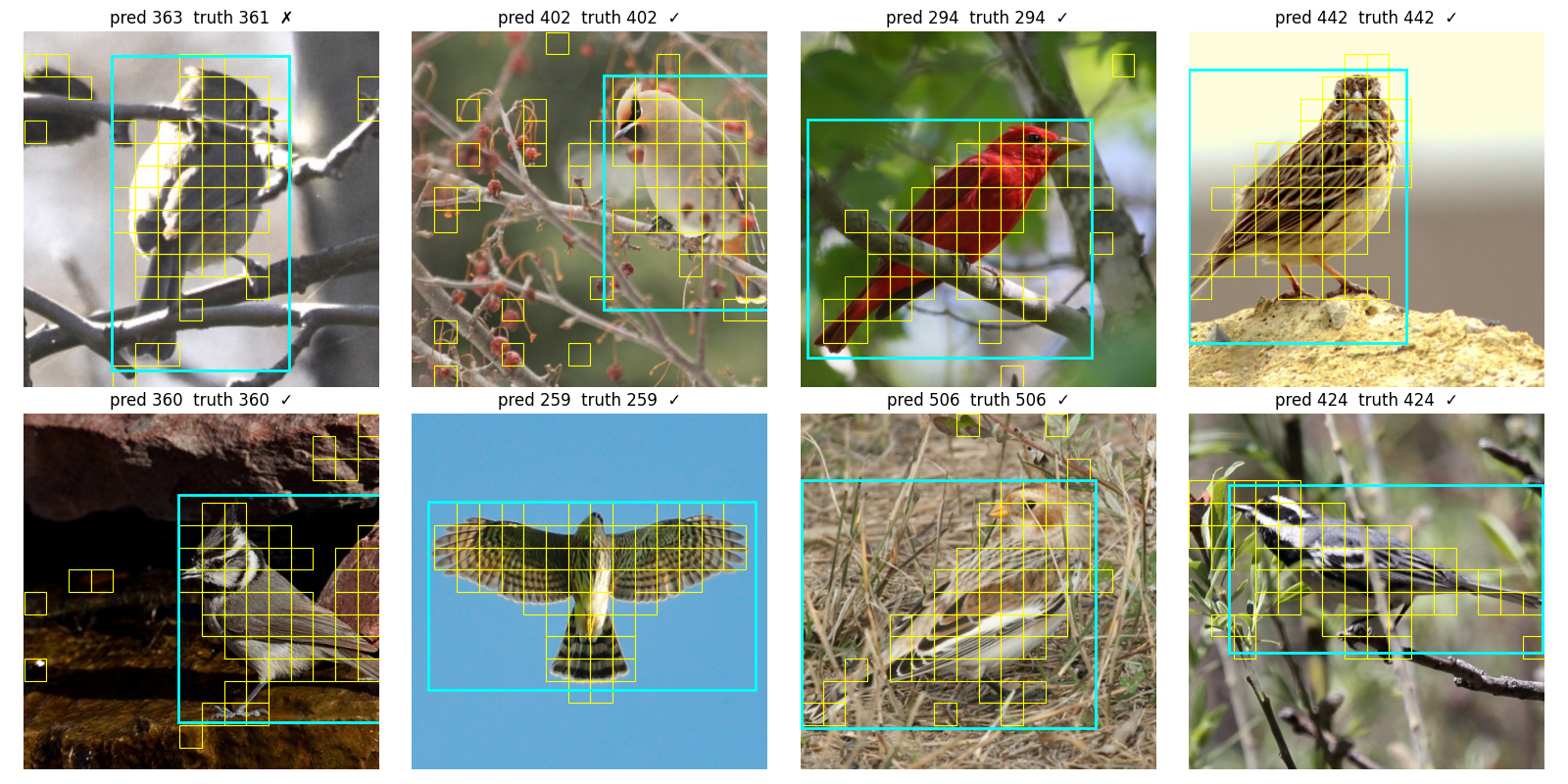}
\end{minipage}\hfill
\begin{minipage}[t]{0.49\textwidth}
\centering\small Aircraft\par
\includegraphics[width=\linewidth]{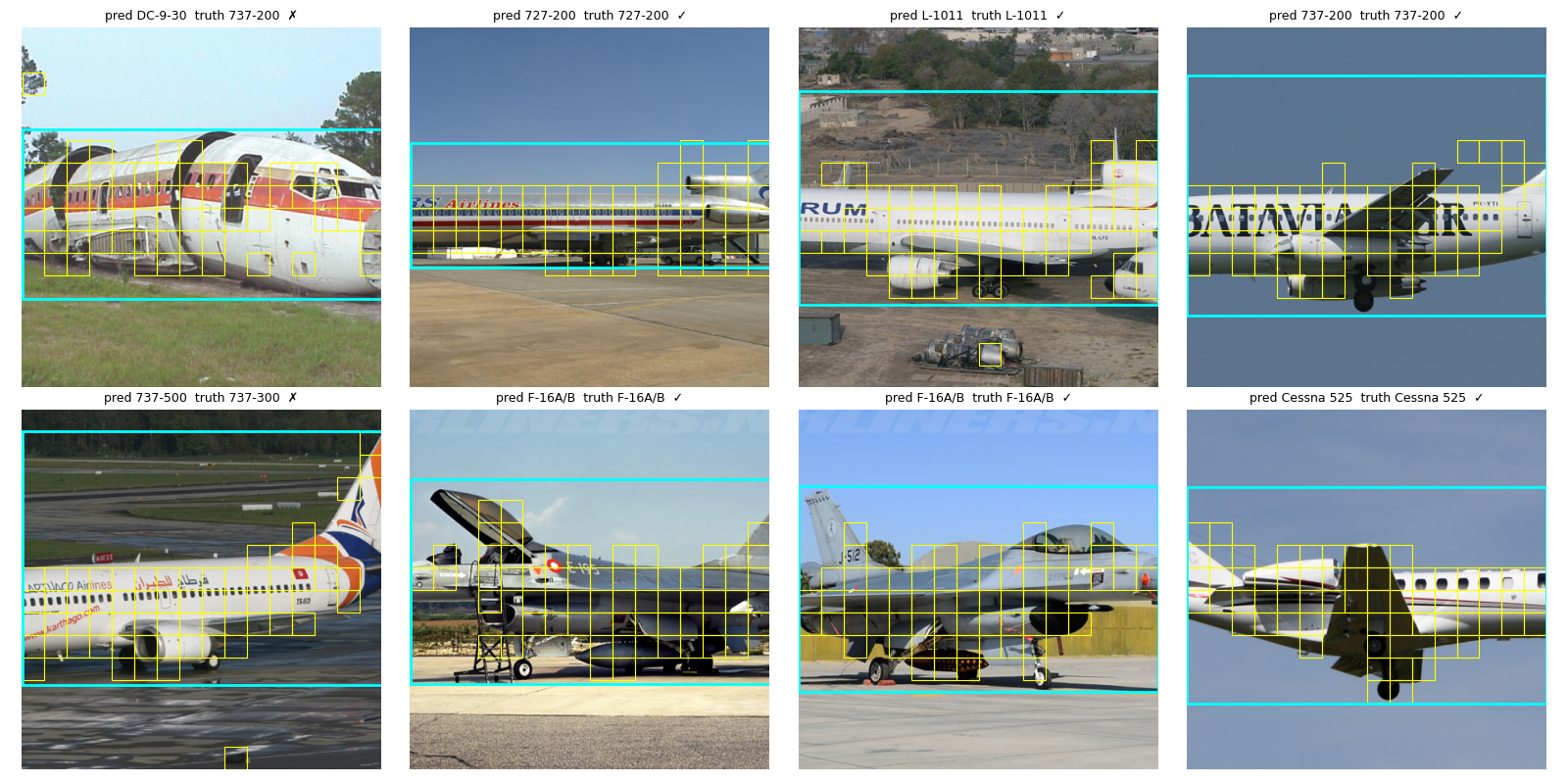}
\end{minipage}
\caption{Unedited selection grids at $K=64$: NABirds (left) and Aircraft (right). Yellow outlines mark the cells FoVT refines; cyan outlines mark annotated object boxes in the displayed crop. Predicted and ground-truth labels are shown as recorded, including errors. Sec.~\ref{sec:where} gives quantitative hit and coverage rates.}
\label{fig:qual}
\end{figure*}

\subsection{Stanford Dogs: Where Looking Closer Does Not Pay}
\label{sec:dogs}
Dogs illustrates a limit of the approach in this study. Uniform accuracy is flat from 336 to 392~px (87.05 and 87.07\%) and reaches 87.40\% at 448~px, a non-significant gain. FoVT shows no significant difference from any uniform model of at least 336~px ($-0.18$ to $+0.17$), and its equal-cost gain over uniform 336 is the smallest of the five datasets. Poor localization does not appear to explain this result: attention selection beats random by $+0.94$ points and the center prior, swapped in at inference, by $+0.97$ points, and 93\% of the selected cells fall on the dog. A model trained with the center prior, however, comes within $0.27$ points (not significant), which happens on no other dataset. Refinement improves on the glance by $+0.83$ points, yet at the cost of uniform 336 the gated model ends only $0.17$ points above it.

We see four plausible, untested explanations. Breed identity is carried by body-level cues such as proportions, coat, and build, which are visible at moderate resolution. The subjects fill most of the frame (mean box area 0.71). The source images are ImageNet photographs of modest resolution~\cite{deng2009imagenet}, so the 448-px view adds interpolated rather than new detail. Finally, DINOv2's pretraining data were curated with retrieval against sources that include ImageNet~\cite{oquab2024dinov2}, so breed information may already be well encoded at coarse scale. Coverage of the object does not explain the result: Cars has a lower coverage share (36\% vs.\ 38\%) and still gains.

\subsection{Efficiency in Practice}
\label{sec:latency}
\paragraph{Batch-normalized forward time}
On an A100 (batch 32, BF16, synthetic 448-px inputs already on the GPU, five warmup and twenty timed forwards), FoVT at $K=64$ runs at 0.56~ms/image. This statistic is synchronized elapsed time divided by the number of timed forwards and by batch size 32; it is not individual-request response time. Uniform 448 runs at 0.75~ms/image ($-26\%$ for FoVT), uniform 392 at 0.58 ($-4\%$), and uniform 336 at 0.43 ($+30\%$, at $+14\%$ MACs). At $K=32$ FoVT takes 0.52~ms/image, slower than uniform 336 despite fewer MACs, and the reuse-based variant takes 0.66--0.68~ms/image (the CF-ViT port was not timed). On this hardware, fused attention kernels~\cite{dao2022flashattention} shrink the quadratic term that foveation removes, and the second backbone launch, attention extraction, and token gathering form a fixed overhead. Arithmetic savings therefore translate only partially into wall-clock savings, a known gap between cost indicators and speed~\cite{dehghani2022misnomer}. The gate's savings are not included in these measurements, since they require batch compaction. Fusing selection and gathering into the forward pass is the main engineering target.

\paragraph{Memory and training cost}
FoVT's peak training memory is 4.64~GiB, compared with 6.35~GiB for uniform 448 ($-27\%$) and 4.98~GiB for uniform 392. In our unoptimized implementation on the A100, a FoVT training epoch takes 1.12--1.19$\times$ as long as a uniform 448-px epoch, because each step runs two forward and backward passes.

\paragraph{Static shapes}
At a fixed $K$, every image yields the same sequence length, so batches need no padding. All refined examples retain this token length at fixed $K$, but the number of refined examples in an incoming batch can vary with gate decisions. The rank rule fixes the total refined count over its evaluated pool, not the subset size in each incoming batch.

\begin{table}[!t]
\centering
\caption{Separate single-seed comparison suite (accuracy \% / transformer GMACs). This suite uses its own splits (Aircraft trains on the 3,334-image \emph{train} split only and selects checkpoints on validation), so its numbers are not comparable to Tables~\ref{tab:main}--\ref{tab:budgets}. ToMe uses the 392-px CUB and 448-px Aircraft/Food checkpoints with $r=24$ and $r=64$. MSViT costs average active tokens and exclude padding.}
\label{tab:external}
\begin{adjustbox}{max width=\columnwidth}
\footnotesize
\setlength{\tabcolsep}{2.5pt}
\begin{tabular}{lccc}
\toprule
Method & CUB & Aircraft & Food-101\\
\midrule
Uniform 448 & 90.01 / 31.61 & 78.61 / 31.61 & 93.06 / 31.61\\
FoVT random & 88.99 / 17.68 & 77.77 / 17.68 & 92.48 / 17.68\\
FoVT attention & 89.99 / 17.68 & 79.93 / 17.68 & 92.92 / 17.68\\
ToMe~\cite{bolya2023tome} & 89.92 / 17.65 & 78.28 / 18.14 & 92.48 / 18.14\\
EViT~\cite{liang2022evit}, keep 0.63 & 89.70 / 17.71 & 78.43 / 17.71 & 92.12 / 17.71\\
EViT~\cite{liang2022evit}, keep 0.70 & 90.09 / 19.56 & 79.30 / 19.56 & 92.48 / 19.56\\
Quadformer~\cite{ronen2023quadformer} & 89.99 / 17.53 & 79.15 / 17.53 & 92.69 / 17.53\\
MSViT~\cite{havtorn2023msvit} & 89.16 / 18.43 & 75.22 / 18.79 & 92.38 / 20.27\\
\bottomrule
\end{tabular}
\end{adjustbox}
\end{table}

\subsection{Token-Reduction and Mixed-Resolution Baselines}
\label{sec:external}
Table~\ref{tab:external} reports a separate, earlier single-seed suite on CUB, Aircraft, and Food-101~\cite{bossard2014food}. It compares FoVT with adaptations of ToMe~\cite{bolya2023tome} (applied at inference with prefix tokens protected; we display the sweep point nearest FoVT's cost), EViT~\cite{liang2022evit} (fine-tuned with fusion at blocks 4, 7, and 10), Quadformer~\cite{ronen2023quadformer} (official quadtree tokenizer and frozen scorer, 640 tokens, learned positions), and MSViT~\cite{havtorn2023msvit} (official batch-shaping loss, learned gate). In that suite FoVT was the most accurate method near its cost on Aircraft and Food-101 and tied Quadformer on CUB. On Food-101 its gains over both EViT settings, MSViT, and ToMe survived Holm correction; Quadformer, which relies on an external scorer, was not significantly different. MSViT's batch-padded costs were close to uniform 448 (29.2--30.8 GMACs). On an RTX 4090 Laptop GPU, FoVT on CUB ran at 1.49~ms/image, compared with 1.87 for uniform 392. These results position FoVT among existing efficient designs but were not repeated under the two-seed protocol.

\section{Limitations}
\label{sec:limitations}
We use one backbone (ViT-S/14) and two seeds. Seed standard deviations are therefore coarse, and our intervals capture test-set sampling but not training variance beyond two runs~\cite{bouthillier2021variance}. The reuse learning-rate controls use a single seed. Our CF-ViT is a port of the official code to a different backbone and training recipe, with its last reuse layer zero-initialized and trained at the backbone learning rate; it was trained on a different GPU type, compared at $\alpha=0.25$ only under the two-seed protocol, and not tuned further. The rank gate needs the test pool; the deployable threshold gate realizes costs up to 2.7\% above target. MAC counts exclude selection, gathering, and patch embedding, and latency was measured on one GPU type without batch compaction for the gate. The training-matched baseline exists only at 448~px, so the equal-cost comparison with uniform 336 includes the glance-loss training effect. The dataset-level correlations rest on five datasets, and the validation splits were too small to make marginal deployment decisions reliably.

\section{Conclusion}
We used a simple foveated ViT to ask when looking closer pays off in fine-grained recognition. The backbone's last-block attention selects useful regions in this setting. With high-resolution input embeddings in the second pass, FoVT shows no Holm-corrected deficit relative to the official CF-ViT port at equal cost. Its small observed advantages do not survive correction in both seeds on any dataset. Feature reuse gave no significant benefit and reduced accuracy when trained at the head learning rate. A confidence gate on the glance skips refinement for the 19--24\% most confident images with mean accuracy unchanged to two decimals in these evaluated runs, and held-out calibration keeps the realized cost within 2.7\% of the target. FoVT runs at $0.56\times$ the cost of a training-matched 448-px model, with no significant accuracy difference on four of five datasets. Its advantage over standard training comes mainly from an auxiliary glance loss that also improves uniform models. The results suggest that the approach benefits tasks where discriminative evidence is small and resolution still helps beyond 336~px, as on NABirds and FGVC-Aircraft, and it brings no significant gain on Stanford Dogs. Future work should replicate these findings with larger backbones and more seeds, train across budgets to remove the Aircraft budget sensitivity, measure gated latency with batch compaction, and test reliable, label-free predictors of the benefit before deployment.

\bibliographystyle{IEEEtran}
\bibliography{refs}

\end{document}